\documentclass{article}
\usepackage{amssymb}
\usepackage{amsfonts}
\usepackage{paralist}
\usepackage{subfigure}
\usepackage{amsmath}
\usepackage{amsthm}
\usepackage{mathtools}
\usepackage{mathrsfs}
\usepackage{soul,color}

\usepackage[round]{natbib}
\usepackage[colorlinks=true,
            linkcolor=blue,
            urlcolor=blue,
            citecolor=blue]{hyperref}
\usepackage[nameinlink]{cleveref}
\usepackage{pgf}
\usepackage[singlespacing]{setspace}
\usepackage{bm, bbm}
\usepackage{times}
\usepackage{pgf,tikz}
\usepackage[utf8]{inputenc}
\usepackage{algorithm}
\usepackage{algorithmic}
\usepackage{enumitem}
\usepackage{array}
\usepackage{float}
\usetikzlibrary{arrows,shapes.arrows,shapes.geometric,shapes.multipart,fit,automata,decorations.pathmorphing,positioning}
\tikzset{
	>=stealth',
	true/.style={
		rectangle,
		draw=black, very thick,
		text width=6.5em,
		minimum height=2em,
		text centered,
		fill=gray, opacity = 0.5},
	punkt/.style={
		rectangle,
		rounded corners,
		draw=black, very thick,
		text width=6.5em,
		minimum height=2em,
		text centered},
	est/.style={
		circle,
		draw=black, very thick,
		text centered},
	shade/.style={
		circle,
		draw=black, very thick, fill=gray!50,
		text centered},
	weight/.style={
		circle,
		draw=black, very thick,
		text width=6.5em,
		minimum height=2em,
		text centered},
	pil/.style={
		->,
		thick,
		shorten <=2pt,
		shorten >=2pt,},
	double/.style={
		<->,
		thick,
		shorten <=2pt,
		shorten >=2pt,},
	dash/.style={
		dashed,
		thick,
		shorten <=2pt,
		shorten >=2pt,},
	dashdouble/.style={
		<->,
		dashed,
		thick,
		shorten <=2pt,
		shorten >=2pt,}
}
\usepackage[resetlabels]{multibib}
\usepackage{appendix}

\def\E{\mathsf{E}}

\def\EIF{\mathsf{EIF}}

\def\argmin{\mathsf{arg \ min}}

\def\calH{\mathcal{H}}

\def\calF{\mathcal{F}}

\def\[{\left[}
\def\]{\right]}
\def\({\left(}
\def\){\right)}
\def\diff{\mathrm{d}}

\usepackage{accents}
\newcommand{\ubar}[1]{\underaccent{\bar}{#1}}
\newcites{SM}{References for Appendix}
\makeatletter
\newenvironment{breakablealgorithm}
  {
   \begin{center}
     \refstepcounter{algorithm}
     \hrule height.8pt depth0pt \kern2pt
     \renewcommand{\caption}[2][\relax]{
       {\raggedright\textbf{\fname@algorithm~\thealgorithm} ##2\par}%
       \ifx\relax##1\relax 
         \addcontentsline{loa}{algorithm}{\protect\numberline{\thealgorithm}##2}%
       \else 
         \addcontentsline{loa}{algorithm}{\protect\numberline{\thealgorithm}##1}%
       \fi
       \kern2pt\hrule\kern2pt
     }
  }{
     \kern2pt\hrule\relax
   \end{center}
  }
\makeatother

\newtheorem{theorem}{Theorem}

\newtheorem{lemma}[theorem]{Lemma}

\newtheorem{proposition}[theorem]{Proposition}
\newtheorem{remark}[theorem]{Remark}

\newcommand*\samethanks[1][\value{footnote}]{\footnotemark[#1]}

\renewcommand{\hat}{\widehat}
\renewcommand{\tilde}{\widetilde}
\renewcommand{\(}{\left(}
\renewcommand{\)}{\right)}
\renewcommand{\[}{\left[}
\renewcommand{\]}{\right]}

\def\NDE{\mathsf{NDE}}
\def\NIE{\mathsf{NIE}}

\usepackage{microtype}
\usepackage{graphicx}
\usepackage{booktabs} 
\usepackage{todonotes}
\newcounter{todocounter}

\definecolor{ForestGreen}{RGB}{34,139,34}

\usepackage[final]{neurips_2022}

\title{DeepMed: Semiparametric Causal Mediation Analysis with Debiased Deep Learning}

\author{%
  Siqi Xu \\
  Department of Statistics and Actuarial Sciences \\
  University of Hong Kong \\
  Hong Kong SAR, China \\
  \texttt{sqxu@hku.hk}
  \And
  Lin Liu\thanks{Co-corresponding authors, alphabetical order} \\
  Institute of Natural Sciences, MOE-LSC, \\ 
  School of Mathematical Sciences, CMA-Shanghai, \\ 
  and SJTU-Yale Joint Center for Biostatistics and Data Science \\
  Shanghai Jiao Tong University and Shanghai Artificial Intelligence Laboratory \\
  Shanghai, China \\
  \texttt{linliu@sjtu.edu.cn}
  \AND
  Zhonghua Liu\samethanks[1] \\
  Department of Biostatistics \\
  Columbia University \\
  New York, NY, USA \\
  \texttt{zl2509@cumc.columbia.edu} \\
}

\date{\today}

\begin{document}

\maketitle

\begin{abstract}
Causal mediation analysis can unpack the black box of causality and is therefore a powerful tool for disentangling causal pathways in biomedical and social sciences, and also for evaluating machine learning fairness. To reduce bias for estimating Natural Direct and Indirect Effects in mediation analysis, we propose a new method called $\mathsf{DeepMed}$ that uses deep neural networks (DNNs) to cross-fit the infinite-dimensional nuisance functions in the efficient influence functions. We obtain novel theoretical results that our $\mathsf{DeepMed}$ method (1) can achieve semiparametric efficiency bound without imposing sparsity constraints on the DNN architecture and (2) can adapt to certain low-dimensional structures of the nuisance functions, significantly advancing the existing literature on DNN-based semiparametric causal inference. Extensive synthetic experiments are conducted to support our findings and also expose the gap between theory and practice. As a proof of concept, we apply $\mathsf{DeepMed}$ to analyze two real datasets on machine learning fairness and reach conclusions consistent with previous findings.
\end{abstract}

\allowdisplaybreaks
\section{Introduction}
\label{sec:intro}

Tremendous progress has been made in this decade on deploying deep neural networks (DNNs) in real-world problems \citep{krizhevsky2012imagenet, wolf2019huggingface, jumper2021highly, brown2022deep}. Causal inference is no exception. 
In semiparametric causal inference, a series of seminal works \citep{chen2020causal, chernozhukov2020adversarial, farrell2021deep} initiated the investigation of statistical properties of causal effect estimators when the nuisance functions (the outcome regressions and propensity scores) are estimated by DNNs. However, there are a few limitations in the current literature that need to be addressed before the theoretical results can be used to guide practice:

\ \ \ \ (1) Most recent works mainly focus on total effect \citep{chen2020causal, farrell2021deep}. In many settings, however, more intricate causal parameters are often of greater interests. In biomedical and social sciences, one is often interested in ``mediation analysis'' to decompose the total effect into direct and indirect effect to unpack the underlying black-box causal mechanism \citep{baron1986moderator}. More recently, mediation analysis also percolated into machine learning fairness. 
For instance, in the context of predicting the recidivism risk, \citet{nabi2018fair} argued that, for a ``fair'' algorithm, sensitive features such as race should have no direct effect on the predicted recidivism risk. If such direct effects can be accurately estimated, one can detect the potential unfairness of a machine learning algorithm. We will revisit such applications in Section \ref{sec:real} and Appendix \ref{app:real}.

\ \ \ \ (2) Statistical properties of DNN-based causal estimators in recent works mostly follow from several (recent) results on the convergence rates of DNN-based nonparametric regression estimators \citep{suzuki2019adaptivity, schmidt2020nonparametric, tsuji2021estimation}, with the limitation of relying on {\it sparse} DNN architectures. The theoretical properties are in turn evaluated by relatively simple synthetic experiments not designed to generate nearly {\it infinite-dimensional} nuisance functions, a setting considered by almost all the above related works. 

The above limitations raise the tantalizing question whether the available statistical guarantees for DNN-based causal inference have practical relevance. In this work, we plan to partially fill these gaps by developing a new method called $\mathsf{DeepMed}$ for semiparametric mediation analysis with DNNs. We focus on the {\it Natural Direct/Indirect Effects} (NDE/NIE) \citep{robins1992identifiability, pearl2001direct} (defined in Section \ref{sec:defn}), but our results can also be applied to more general settings; see Remark \ref{rem:dr}. The $\mathsf{DeepMed}$ estimators leverage the ``multiply-robust'' property of the efficient influence function (EIF) of NDE/NIE \citep{tchetgen2012semiparametric, farbmacher2022causal} (see Proposition \ref{thm:master} in Section \ref{sec:semi}), together with the flexibility and superior predictive power of DNNs (see Section \ref{sec:deepmed} and Algorithm \ref{alg1}). 
In particular, we also make the following novel contributions to deepen our understanding of DNN-based semiparametric causal inference:
\begin{itemize}
    \item On the theoretical side, we obtain new results that our $\mathsf{DeepMed}$ method can achieve semiparametric efficiency bound without imposing sparsity constraints on the DNN architecture and can adapt to certain low-dimensional structures of the nuisance functions (see Section \ref{sec:stat}), thus significantly advancing the existing literature on DNN-based semiparametric causal inference. {\it Non-sparse} DNN architecture is more commonly employed in practice \citep{farrell2021deep}, and the low-dimensional structures of nuisance functions can help avoid curse-of-dimensionality. These two points, taken together, significantly advance our understanding of the statistical guarantee of DNN-based causal inference.
    \item More importantly, on the empirical side, in Section \ref{sec:sim}, we designed sophisticated  synthetic experiments to simulate nearly {\it infinite-dimensional} functions, which are much more complex than those in previous related works \citep{chen2020causal, farrell2021deep, adcock2021gap}. We emphasize that these nontrivial experiments could be of independent interest to the theory of deep learning beyond causal inference, to further expose the gap between deep learning theory and practice \citep{adcock2021gap, gottschling2020troublesome}; see Remark \ref{beyond} for an extended discussion. As a proof of concept, in Section \ref{sec:real} and Appendix \ref{app:real}, we also apply $\mathsf{DeepMed}$ to re-analyze two real-world datasets on algorithmic fairness and reach similar conclusions to related works.
    \item Finally, a user-friendly R package can be found at \href{https://github.com/siqixu/DeepMed}{https://github.com/siqixu/DeepMed}. Making such resources available helps enhance reproducibility, a highly recognized problem in all scientific disciplines, including (causal) machine learning \citep{pineau2021improving, kaddour2022causal}.
\end{itemize}

\section{Definition, identification, and estimation of NDE and NIE}
\label{sec:mediation}

\subsection{Definition of NDE and NIE}
\label{sec:defn}

Throughout this paper, we denote $Y$ as the primary outcome of interest, $D$ as a binary treatment variable, $M$ as the mediator on the causal pathway from $D$ to $Y$, and $X \in [0, 1]^{p}$ (or more generally, compactly supported in $\mathbb{R}^{p}$) as baseline covariates including all potential confounders. We denote the observed data vector as $O \equiv (X, D, M, Y)$. Let $M (d)$ denote the potential outcome for the mediator when setting $D = d$ and $Y (d, m)$ be the potential outcome of $Y$ under $D = d$ and $M = m$, where $d \in \{0, 1\}$ and $m$ is in the support $\mathcal{M}$ of $M$. We define the average total (treatment) effect as $\tau_{tot} \coloneqq \E [Y (1, M (1)) - Y (0, M (0))]$, the average NDE of the treatment $D$ on the outcome $Y$ when the mediator takes the natural potential outcome when $D = d$ as $\tau_{\NDE} (d) \coloneqq \E [Y (1, M (d)) - Y (0, M(d))]$, and the average NIE of the treatment $D$ on the outcome $Y$ via the mediator $M$ as $\tau_{\NIE} (d) \coloneqq \E [Y (d, M(1)) - Y (d, M(0))]$. We have the trivial decomposition $\tau_{tot} \equiv \tau_{\NDE} (d) + \tau_{\NIE} (d')$ for $d \neq d'$. In causal mediation analysis, the parameters of interest are $\tau_{\NDE} (d)$ and $\tau_{\NIE} (d)$. 

\subsection{Semiparametric multiply-robust estimators of NDE/NIE}
\label{sec:semi}

Estimating $\tau_{\NDE} (d)$ and $\tau_{\NIE} (d)$ can be reduced to estimating $\phi (d, d') \coloneqq \E [Y (d, M (d'))]$ for  $d, d' \in \{0, 1\}$. We make the following standard identification assumptions:
\begin{itemize}
	\item[i.] \textbf{Consistency}: if $D = d$, then $M = M (d)$ for all $d \in \{0, 1\}$; while if $D = d$ and $M = m$, then
	$Y = Y(d, m)$ for all $d \in \{0, 1\}$ and all $m$ in the support of $M$.
	\item[ii.] \textbf{Ignorability}: $Y (d, m) \perp D | X$, $Y (d, m) \perp M | X, D$, $M (d) \perp D | X$, and $Y (d, m) \perp M (d') | X$, almost surely for all $d, \in \{0, 1\}$ and all $m \in \mathcal{M}$. The first three conditions are, respectively, no unmeasured treatment-outcome, mediator-outcome and treatment-mediator confounding, whereas the fourth condition is often referred to as the ``cross-world'' condition. We provide more detailed comments on these four conditions in Appendix \ref{app:ignore}.
	\item[iii.] \textbf{Positivity}: The propensity score $a (d | X) \equiv \mathsf{Pr} (D = d | X) \in (c, C)$ for some constants $0 < c \leq C < 1$, almost surely for all $d \in\{0, 1\}$; $f (m | X, d)$, the conditional density (mass) function of $M = m$ (when $M$ is discrete) given $X$ and $D = d$, is strictly bounded between $[\ubar{\rho}, \bar{\rho}]$ for some constants $0 < \ubar{\rho} \leq \bar{\rho} < \infty$ almost surely for all $m$ in $\mathcal{M}$ and all $d \in \{0, 1\}$.
\end{itemize}
Under the above assumptions, the causal parameter $\phi (d, d')$ for $d, d' \in \{0, 1\}$ can be identified as either of the following three observed-data functionals:
\small
\begin{equation} \label{eq:id}
\begin{split}
\phi (d, d') & \equiv \E \left[ \frac{\mathbbm{1} \{D = d\} f (M | X, d')Y}{a (d | X) f (M | X, d)} \right] \equiv \E \left[ \frac{\mathbbm{1} \{D = d'\}}{a (d' | X)} \mu (X, d, M) \right] \\
& \equiv \int \mu (x, d, m) f (m | x, d') p (x) \ \diff m \diff x,
\end{split}
\end{equation}
\normalsize
where $\mathbbm{1} \{\cdot\}$ denotes the indicator function, $p (x)$ denotes the marginal density of $X$, and $\mu (x, d, m) \coloneqq \E [Y | X = x, D = d, M = m]$ is the outcome regression model, for which we also make the following standard boundedness assumption:
\begin{itemize}
\item[iv.] $\mu (x, d, m)$ is also strictly bounded between $[-R, R]$ for some constant $R > 0$.
\end{itemize}
Following the convention in the semiparametric causal inference literature, we call $a, f, \mu$ ``nuisance functions''. \citet{tchetgen2012semiparametric} derived the EIF of $\phi (d, d')$: $\mathsf{EIF}_{d, d'} \equiv \psi_{d, d'} (O) - \phi (d, d')$, where
{\small
\begin{align}
\psi_{d, d'} (O) & = \frac{\mathbbm{1} \{D = d\} \cdot f (M | X, d')}{a (d | X) \cdot f (M | X, d)} (Y - \mu (X, d, M)) \notag \\
& + \left( 1 - \frac{\mathbbm{1} \{D = d'\}}{a (d' | X)} \right) \int_{m \in \mathcal{M}} \mu (X, d, m) f (m | X, d') \diff m + \frac{\mathbbm{1} \{D = d'\}}{a (d' | X)} \mu (X, d, M). \label{eq6}
\end{align}
}

The nuisance functions $\mu(x, d, m)$, $a (d | x)$ and $f (m | x, d)$ appeared in $\psi_{d, d'} (o)$ are unknown and generally high-dimensional. But with a sample $\mathcal{D} \equiv \{O_{j}\}_{j = 1}^{N}$ of the observed data, based on $\psi_{d, d'} (o)$, one can construct the following generic sample-splitting multiply-robust estimator of $\phi (d, d')$:
\begin{equation}\label{mr}
\tilde{\phi} (d, d') = \frac{1}{n} \sum_{i \in \mathcal{D}_{n}} \tilde{\psi}_{d, d'} (O_{i}),
\end{equation}
where $\mathcal{D}_{n} \equiv \{O_{i}\}_{i = 1}^{n}$ is a subset of all $N$ data, and $\tilde{\psi}_{d, d'} (o)$ replaces the unknown nuisance functions $a, f, \mu$ in $\psi_{d, d'} (o)$ by some generic estimators $\tilde{a}, \tilde{f}, \tilde{\mu}$ computed using the remaining $N - n$ {\it nuisance sample data}, denoted as $\mathcal{D}_{\nu}$. Cross-fit is then needed to recover the information lost due to sample splitting; see Algorithm \ref{alg1}. It is clear from \eqref{eq6} that $\tilde{\phi} (d, d')$ is a consistent estimator of $\phi (d, d')$ as long as any two of $\tilde{a}, \tilde{f}, \tilde{\mu}$ are consistent estimators of the corresponding true nuisance functions, hence the name ``multiply-robust''. Throughout this paper, we take $n \asymp N - n$ and assume:
\begin{itemize}
\item[v.] Any nuisance function estimators are strictly bounded within the respective lower and upper bounds of $a, f, \mu$.
\end{itemize}
To further ease notation, we define: for any $d \in \{0, 1\}$, $
    r_{a, d} \coloneqq \left\{ \int \delta_{a, d} (x)^{2} \diff F (x) \right\}^{1 / 2}, r_{f, d} \coloneqq \left\{ \int \delta_{f, d} (x, m)^{2} \diff F (x, m | d = 0) \right\}^{1 / 2},
$ and $
    r_{\mu, d} \coloneqq \left\{ \int \delta_{\mu, d} (x, m)^{2} \diff F (x, m | d = 0) \right\}^{1 / 2},
$
where $\delta_{a, d} (x) \coloneqq \tilde{a} (d | x) - a (d | x)$, $\delta_{f, d} (x, m) \coloneqq \tilde{f} (m | x, d) - f (m | x, d)$ and $\delta_{\mu, d} (x, m) \coloneqq \tilde{\mu} (x, d, m) - \mu (x, d, m)$ are point-wise estimation errors of the estimated nuisance functions. In defining the above $L_{2}$-estimation errors, we choose to take expectation with respect to (w.r.t.) the law $F (m, x | d = 0)$ {\it only} for convenience, with no loss of generality by Assumptions iii and v.

To show the cross-fit version of $\tilde{\phi} (d, d')$ is semiparametric efficient for $\phi (d, d')$, we shall demonstrate under what conditions $\sqrt{n} (\tilde{\phi} (d, d') - \phi (d, d')) \overset{\mathcal{L}}{\rightarrow} \mathcal{N} (0, \E [\EIF_{d, d'}^{2}])$ \citep{newey1990semiparametric}. The following proposition on the statistical properties of $\tilde{\phi} (d, d')$ is a key step towards this objective.

\begin{proposition}\label{thm:master}
Denote $\mathrm{Bias} (\tilde{\phi} (d, d')) \coloneqq \E [\tilde{\phi} (d, d') - \phi (d, d') | \mathcal{D}_{\nu}]$ as the bias of $\tilde{\phi} (d, d')$ conditional on the nuisance sample $\mathcal{D}_{\nu}$. Under Assumptions i -- v, $\mathrm{Bias} (\tilde{\phi} (d, d'))$ is of second-order:
\begin{align}
|\mathrm{Bias} (\tilde{\phi} (d, d'))| \lesssim \max \left\{ r_{a, d} \cdot r_{f, d}, \underset{d'' \in \{0, 1\}}{\max} r_{f, d''} \cdot r_{\mu, d}, r_{a, d} \cdot r_{\mu, d} \right\}. \label{l2}
\end{align}
Furthermore, if the RHS of \eqref{l2} is $o (n^{- 1 / 2})$, then
\begin{align}
\sqrt{n} \left( \tilde{\phi} (d, d') - \phi (d, d') \right) = \frac{1}{\sqrt{n}} \sum_{i = 1}^{n} (\psi_{d, d'} (O_{i}) - \phi (d, d')) + o (1) \overset{d}{\rightarrow} \mathcal{N} \left( 0, \E \left[ \EIF_{d, d'}^{2} \right] \right).  \label{an}
\end{align}
\end{proposition}
Although the above result is a direct consequence of the EIF $\psi_{d, d'} (O)$, we prove Proposition \ref{thm:master} in Appendix \ref{app:bias} for completeness.

\begin{remark}
\label{rem:dr}
The total effect $\tau_{tot} = \phi (1, 1) - \phi (0, 0)$ can be viewed as a special case, for which $d = d'$ for $\phi (d, d')$. Then $\mathsf{EIF}_{d, d} \equiv \mathsf{EIF}_{d}$ corresponds to the nonparametric EIF of $\phi (d, d) \equiv \phi (d) \equiv \E [Y (d, M (d))]$:
\begin{align*}
\mathsf{EIF}_{d} = \psi_{d} (O) - \phi (d) \text{ with } \psi_{d} (O) = \frac{\mathbbm{1} \{D = d\}}{a (d | X)} Y + \left( 1 - \frac{\mathbbm{1} \{D = d\}}{a (d | X)} \right) \mu (X, d),
\end{align*}
where $\mu (x, d) \coloneqq \E [Y | X = x, D = d]$. Hence all the theoretical results in this paper are applicable to total effect estimation. Our framework can also be applied to all the statistical functionals that satisfy a so-called ``mixed-bias'' property, characterized recently in \citet{rotnitzky2021characterization}. This class includes the quadratic functional, which is important for uncertainty quantification in machine learning.
\end{remark}


\section{Estimation and inference of NDE/NIE using DeepMed}
\label{sec:dnn}

We now introduce $\mathsf{DeepMed}$, a method for mediation analysis with nuisance functions estimated by DNNs. By leveraging the second-order bias property of the multiply-robust estimators of NDE/NIE (Proposition \ref{thm:master}), we will derive statistical properties of $\mathsf{DeepMed}$ in this section. The nuisance function estimators by DNNs are denoted as $\hat{a}, \hat{f}, \hat{\mu}$.


\subsection{Details on DeepMed}
\label{sec:deepmed}
First, we introduce the fully-connected feed-forward neural network with the rectified linear units (ReLU) as the activation function for the hidden layer neurons (FNN-ReLU), which will be used to estimate the nuisance functions. Then, we will introduce an estimation procedure using a $V$-fold cross-fitting with sample-splitting to avoid the Donsker-type empirical-process assumption on the nuisance functions, which, in general, is violated in high-dimensional setup. Finally, we provide the asymptotic statistical properties of the DNN-based estimators of $\tau_{tot}$, $\tau_{\NDE} (d)$ and $\tau_{\NIE} (d)$.

We denote the ReLU activation function as $\sigma (u) \coloneqq \max (u, 0)$ for any $u \in \mathbb{R}$. Given vectors $x, b$, we denote $\sigma_{b} (x) \coloneqq \sigma (x - b)$, with $\sigma$ acting on the vector $x - b$ component-wise. 

Let $\mathcal{F}_{\mathrm{nn}}$ denote the class of the FNN-ReLU functions
\begin{equation*}
\mathcal{F}_{\mathrm{nn}} \coloneqq \left\{ f: \mathbb{R}^{p} \rightarrow \mathbb{R}; f (x) = W^{(L)} \sigma_{b^{(L)}} \circ \cdots \circ W^{(1)} \sigma_{b^{(1)}} (x) \right\},
\end{equation*}
where $\circ$ is the composition operator, $L$ is the number of layers (i.e. depth) of the network, and for $l = 1, \cdots, L$, $W^{(l)}$ is a $K_{l + 1} \times K_{l}$-dimensional weight matrix with $K_{l}$ being the number of neurons in the $l$-th layer (i.e. width) of the network, with $K_{1} = p$ and $K_{L + 1} = 1$, and $b^{(l)}$ is a $K_{l}$-dimensional vector. To avoid notation clutter, we concatenate all the network parameters as $\Theta = (W^{(l)}, b^{(l)}, l = 1, \cdots, L)$ and simply take $K_{2} = \cdots = K_{L} = K$. We also assume $\Theta$ to be bounded: $\Vert \Theta \Vert_{\infty} \leq B$ for some universal constant $B > 0$. We may let the dependence on $L$, $K$, $B$ explicit by writing $\calF_{nn}$ as $\mathcal{F}_{\mathrm{nn}} (L, K, B)$.

$\mathsf{DeepMed}$ estimates $\tau_{tot}, \tau_{\NDE} (d), \tau_{\NIE} (d)$ by \eqref{mr}, with the nuisance functions $a, f, \mu$ estimated using $\calF_{nn}$ with the $V$-fold cross-fitting strategy, summarized in Algorithm \ref{alg1} below; also see \citet{farbmacher2022causal}. $\mathsf{DeepMed}$ inputs the observed data $\mathcal{D} \equiv \{O_{i}\}_{i = 1}^{N}$ and outputs the estimated total effect $\hat{\tau}_{tot}$, NDE $\hat{\tau}_{\NDE} (d)$ and NIE $\hat{\tau}_{\NIE} (d)$, together with their variance estimators $\hat{\sigma}_{tot}^2$, $\hat{\sigma}^{2}_{\NDE} (d)$ and $\hat{\sigma}^{2}_{\NIE} (d)$.

{\small\begin{breakablealgorithm}
	\renewcommand{\algorithmicrequire}{\textbf{Input:}}
	\renewcommand{\algorithmicensure}{\textbf{Output:}}
	\caption{$\mathsf{DeepMed}$ with $V$-fold cross-fitting}
	\label{alg1}
	\begin{algorithmic}[1]
		\STATE Choose some integer $V$ (usually $V \in \{2, 3, \cdots, 10\}$)
		\STATE Split the $N$ observations into $V$ subsamples $I_{v} \subset \{1, \cdots, N\} \equiv [N]$ with equal size $n = N / V$; \\
		\FOR{$v = 1, \cdots, V$:}
       		\STATE Fit the nuisance functions by DNNs using observations in $[N] \setminus I_v$
		    \STATE Compute the nuisance functions in the subsample $I_v$ using the estimated DNNs in step 4
		    \STATE Obtain $\{\hat{\psi}_{d} (O_{i}), \hat{\psi}_{d, d'} (O_{i})\}_{i \in I_v}$ for the subsample $I_v$ based on \eqref{eq6}, respectively, with the nuisance functions replaced by their estimates in step 5
		\ENDFOR
		\STATE Estimate average potential outcomes by $\hat{\phi} (d) \coloneqq \frac{1}{N} \underset{i = 1}{\overset{N}{\sum}} \hat{\psi}_{d} (O_{i})$, $\hat{\phi} (d, d') \coloneqq \frac{1}{N} \underset{i = 1}{\overset{N}{\sum}} \hat{\psi}_{d, d'} (O_{i})$
		\STATE Estimate causal effects by $\hat{\tau}_{tot}$, $\hat{\tau}_{\NDE} (d)$ and $\hat{\tau}_{\NIE} (d)$ with $\hat{\phi} (d)$ and $\hat{\phi} (d, d')$
        \STATE Estimate the variances of $\hat{\tau}_{tot}$, $\hat{\tau}_{\NDE} (d)$ and $\hat{\tau}_{\NIE} (d)$ by: \\
        \small $\hat{\sigma}_{tot}^2 \coloneqq \frac{1}{N^2} \underset{i = 1}{\overset{N}{\sum}} (\hat{\psi}_{1} (O_{i}) - \hat{\psi}_{0} (O_{i}))^2 - \frac{1}{N} \hat{\tau}_{tot}^2$; \mbox{$\hat{\sigma}^2_{\NDE} (d) \coloneqq \frac{1}{N^2} \underset{i = 1}{\overset{N}{\sum}} (\hat{\psi}_{1, d} (O_{i}) - \hat{\psi}_{0, d} (O_{i}))^2 - \frac{1}{N} \hat{\tau}_{\NDE}^2 (d)$}; \\
        $\hat{\sigma}^2_{\NIE} (d) \coloneqq \frac{1}{N^2} \underset{i = 1}{\overset{N}{\sum}} (\hat{\psi}_{d, 1} (O_{i}) - \hat{\psi}_{d, 0} (O_{i}))^2 - \frac{1}{N} \hat{\tau}_{\NIE}^2 (d)$
        
	\ENSURE \normalsize $\hat{\tau}_{tot}$, $\hat{\tau}_{\NDE}(d)$, $\hat{\tau}_{\NIE}(d)$, $\hat{\sigma}_{tot}^2$, $\hat{\sigma}^2_{\NDE}(d)$ and $\hat{\sigma}^2_{\NIE}(d)$
	\end{algorithmic}  
\end{breakablealgorithm}}
\begin{remark}[Continuous or multi-dimensional mediators] \label{rem:bayes}
For binary treatment $D$ and continuous or multi-dimensional $M$, to avoid nonparametric/high-dimensional conditional density estimation, we can rewrite $\frac{f (m | x, d')}{a (d | x) f (m | x, d)}$ as $\frac{1 - a (d | x, m)}{a (d | x, m) (1 - a (d | x))}$ by the Bayes' rule and the integral w.r.t. $f (m | x, d')$ in \eqref{eq6} as $\E [\mu (X, d, M) | X = x, D = d']$. Then we can first estimate $\mu (x, d, m)$ by $\hat{\mu} (x, d, m)$ and in turn estimate $\E [\mu (X, d, M) | X = x, D = d']$ by regressing $\hat{\mu} (X, d, M)$ against $(X, D)$ using the FNN-ReLU class. We mainly consider binary $M$ to avoid unnecessary complications; but see Appendix \ref{app:real} for an example in which this strategy is used. Finally, the potential incompatibility between models posited for $a (d | x)$ and $a (d | x, m)$ and the joint distribution of $(X, A, M, Y)$ is not of great concern under the semiparametric framework because all nuisance functions are estimated nonparametrically; again, see Appendix \ref{app:real} for an extended discussion.
\end{remark}

\subsection{Statistical properties of DeepMed: Non-sparse DNN architecture and low-dimensional structures of the nuisance functions}
\label{sec:stat}

According to Proposition \ref{thm:master}, to analyze the statistical properties $\mathsf{DeepMed}$, it is sufficient to control the $L_{2}$-estimation errors of nuisance function estimates $\hat{a}, \hat{f}, \hat{\mu}$ fit by DNNs. To ease presentation, we first study the theoretical guarantees on the $L_{2}$-estimation error for a generic nuisance function $g: W \in [0, 1]^{p} \rightarrow Z \in \mathbb{R}$, for which we assume:
\begin{itemize}
    \item[vi.] $Z = g (W) + \xi$, with $\xi$ sub-Gaussian with mean zero and independent of $W$.
\end{itemize}
Note that when $g$ corresponds to $a, f, \mu$, $(W, Z)$ corresponds to $(X, \mathbbm{1} (D = 1))$, $((X, D), \mathbbm{1} (M = 1))$ and $((X, D, M), Y)$, respectively.

We denote the DNN output from the nuisance sample $\mathcal{D}_{\nu}$ as $\hat{g}$. For theoretical results, we consider $\hat{g}$ as the following empirical risk minimizer (ERM):
\begin{align}\label{erm}
\hat{g} \coloneqq \underset{\bar{g} \in \mathcal{F}_{\mathrm{nn}} (L, K, B)}{\argmin} \sum_{i \in \mathcal{D}_{\nu}} \left( Z_{i} - \bar{g} (W_{i}) \right)^{2}.
\end{align}

To avoid model misspecification, one often assumes $g \in \mathcal{G}$, where $\mathcal{G}$ is some {\it infinite-dimensional} function space. A common choice is $\mathcal{G} = \mathcal{H}_{p} (\alpha; C)$, the H\"{o}lder ball on the input domain $[0, 1]^{p}$, with smoothness exponent $\alpha$ and radius $C$. H\"{o}lder space is one of the most well-studied function spaces in statistics and it is convenient to quantify its complexity by a single smoothness parameter $\alpha$; see Appendix \ref{app:holder} for a review. It is well-known that estimating H\"{o}lder functions suffers from curse-of-dimensionality \citep{stone1982optimal}. One remedy is to consider the following generalized H\"{o}lder space, by imposing certain low-dimensional structures on $g$:
\begin{align*}
\calH_{k}^{\dag} (\alpha; C) \coloneqq \left\{ g (w) = h (\Gamma w): h \in \calH_{k} (\alpha; C), \Gamma \in \mathbb{R}^{k \times p} \text{ unknown}, k \leq p \right\}.
\end{align*}
\begin{remark}
The above definition contains $g (w) = h (w_{I})$, where $I \subset \{1, \cdots, p\}$, as a special case, in which $g$ is assumed to only depend on a subset of the feature vector $w$. One can easily generalize the above definition to additive models $g (w) = \sum_{j = 1}^{p} h_{j} (w_{j})$ where $h_{j} \in \calH_{k_{j}} (\alpha_{j}; C_{j})$, allowing even more modeling flexibility. To avoid complications, we only consider the above simpler model.
\end{remark}

We can show that the ERM estimator $\hat{g}$ \eqref{erm} from the FNN-ReLU class $\mathcal{F}_{\mathrm{nn}} (L, K, B)$ attains the optimal estimation rate over $\calH_{k}^{\dag} (\alpha; C)$ up to log factors, by choosing the depth and width appropriately without assuming sparse neural nets.
\begin{lemma}\label{lem:holder}
Under Assumptions iii -- vi, if $g \in \calH_{k}^{\dag} (\alpha; C)$ for $k \leq p$, with $L K \asymp n^{\frac{k}{2 (k + 2 \alpha)}}$, we have 
$\sup_{g \in \calH_{k}^{\dag} (\alpha; C)} \left\{ \E \left[ (g (W) - \hat{g} (W))^{2} \right] \right\}^{1 / 2} \lesssim n^{- \frac{\alpha}{2 \alpha + k}} (\log n)^{3}$.
\end{lemma}

Lemma \ref{lem:holder}, together with Proposition \ref{thm:master}, implies the main theoretical result of the paper.
\begin{theorem}\label{thm:holder}
Under Assumptions i -- vi and the following condition on $a, f, \mu$: $a \in \calH_{k}^{\dag} (\alpha_{a}; C), f \in \calH_{k}^{\dag} (\alpha_{f}; C), \mu \in \calH_{k}^{\dag} (\alpha_{\mu}; C)$, with
{\small\begin{align} \label{rate}
\min \left\{ \frac{\alpha_{a}}{2 \alpha_{a} + k} + \frac{\alpha_{f}}{2 \alpha_{f} + k}, \frac{\alpha_{f}}{2 \alpha_{f} + k} + \frac{\alpha_{\mu}}{2 \alpha_{\mu} + k}, \frac{\alpha_{a}}{2 \alpha_{a} + k} + \frac{\alpha_{\mu}}{2 \alpha_{\mu} + k} \right\} > \frac{1}{2} + \epsilon,
\end{align}}
for $k \leq p$ and some arbitrarily small $\epsilon > 0$, if ~ $\hat{a}$, $\hat{f}$, $\hat{\mu}$ are respectively the ERM \eqref{erm} from FNN-ReLU classes $\calF_{nn} (L_{a}, K_{a}, B)$, $\calF_{nn} (L_{f}, K_{f}, B)$, $\calF_{nn} (L_{\mu}, K_{\mu}, B)$, of which the product of the depth and width satisfies $L_{g} K_{g} \asymp n^{\frac{k}{2 (k + 2 \alpha_{g})}}$ for $g \in \{a, f, \mu\}$, then the $\mathsf{DeepMed}$ estimators $\hat{\tau}_{tot}$, $\hat{\tau}_{\NDE}(d)$ and $\hat{\tau}_{\NIE}(d)$ computed by Algorithm \ref{alg1} are semiparametric efficient:
\begin{align*}
\hat{\sigma}^{-1}_{tot} (\hat{\tau}_{tot} - \tau_{tot}), \hat{\sigma}^{-1}_{\NDE} (d) (\hat{\tau}_{\NDE} (d) - \tau_{\NDE} (d)), \hat{\sigma}_{\NIE}^{-1} (d) (\hat{\tau}_{\NIE} (d) - \tau_{\NIE} (d)) \stackrel{\mathcal{L}}{\longrightarrow} \mathcal{N} (0, 1), \text{ with}
\end{align*}
$N \hat{\sigma}_{tot}^{2} \stackrel{p}{\rightarrow} \E [(\EIF_{1} - \EIF_{0})^{2}]$, $N \hat{\sigma}_{\NDE}^{2} (d) \stackrel{p}{\rightarrow} \E [(\EIF_{1, d} - \EIF_{0, d})^{2}]$, and $N \hat{\sigma}_{\NIE}^{2} (d) \stackrel{p}{\rightarrow} \E [(\EIF_{d, 1} - \EIF_{d, 0})^{2}]$, i.e. $\hat{\sigma}_{tot}^{2}$, $\hat{\sigma}_{\NDE}^{2}$ and $\hat{\sigma}_{\NIE}^{2}$ are consistent variance estimators.
\end{theorem}

\begin{remark}
To unload notation in the above theorem, consider the special case where the smoothness of all the nuisance functions coincides, i.e. $\alpha_{a} = \alpha_{f} = \alpha_{\mu} = \alpha$. Then Condition \eqref{rate} reduces to $\alpha > k / 2 + \epsilon$ for some arbitrarily small $\epsilon > 0$. For example, if the covariates $X$ have dimension $p = 2$ and no low-dimensional structures are imposed on the nuisance functions (i.e. $k \equiv p$), one needs $\alpha > 1$ to ensure semiparametric efficiency of the $\mathsf{DeepMed}$ estimators.
\end{remark}

We emphasize that Lemma \ref{lem:holder} and Theorem \ref{thm:holder} do not constrain the network sparsity $S$, better reflecting how DNNs are usually used in practice. Theorem \ref{thm:holder} advances results on total and decomposition effect estimation with non-sparse DNNs \citep[Theorem 1]{farrell2021deep} in terms of (1) weaker smoothness conditions and (2) adapting to certain low-dimensional structures of the nuisance functions. The proof of Lemma \ref{lem:holder} follows from a combination of the improved DNN approximation rate obtained in \citet{lu2021deep, jiao2021deep} and standard DNN metric entropy bound \citep{suzuki2019adaptivity}. We prove Lemma \ref{lem:holder} and Theorem \ref{thm:holder} in Appendix \ref{app:holder} for completeness. One weakness of Lemma \ref{lem:holder} and Theorem \ref{thm:holder}, as well as in other contemporary works \citep{chen2020causal, farrell2021deep}, is the lack of algorithmic/training process considerations \citep{chen2022learning}; see Remark \ref{fp} and Appendix \ref{app:sim} for extended discussions.

\begin{remark}[Explicit input-layer regularization]
Training DNNs in practice involves hyperparameter tuning, including the depth $L$ and width $K$ in Theorem \ref{thm:holder} and others like epochs. In the synthetic experiments, we consider the nuisance functions only depending on a $k$-subset of $p$-dimensional input. A reasonable heuristic is to add $L_{1}$-regularization in the input-layer of the DNN. Then the regularization weight $\lambda$ is also a hyperparameter. In practice, we simply use cross-validation to select the hyperparameters that minimize the validation loss. We leave its theoretical justification and the performance of other alternative approaches such as the minimax criterion \citep{robins2020double, cui2019selective} to future works.
\end{remark}

\section{Synthetic experiments}
\label{sec:sim}
In this section and Appendix \ref{app:sim}, we showcase five synthetic experiments. Since ground truth is rarely known in real data, we believe synthetic experiments play an equally, if not more, important role as real data. Before describing the experimental setups, we garner the following key take-home message:
\vspace{-0.2em}
\begin{itemize}
\setlength\itemsep{0.01cm}
\item[(a)] Compared with the other competing methods, $\mathsf{DeepMed}$ exhibits better finite-sample performance in most of our experiments;
\item[(b)] Cross-validation for DNN hyperparameter tuning works reasonably well in our experiments;
\item[(c)] We find $\mathsf{DeepMed}$ with explicit regularization in the input layer improves performance (see Table \ref{Table: no regularization}) when the true nuisance functions have certain low-dimensional structures in their dependence on the covariates. \citet{farrell2021deep} warned against blind explicit regularization in DNNs for total effect estimation. Our observation does not contradict \citet{farrell2021deep} as (1) the purpose of the input-layer regularization is not to control the sparsity of the DNN architecture and (2) we do not further regularize hidden layers;
\item[(d)] Experimental setups for Cases 3 to 5 generate nuisance functions that are nearly {\it infinite-dimensional} and close to the boundary of a H\"{o}lder ball with a given smoothness exponent \citep{liu2020nearly, li2005robust}. Thus these synthetic experiments should be better benchmarks than Cases 1 and 2 or settings in other related works such as \citet{farrell2021deep}. We hope that these highly nontrivial synthetic experiments are helpful to researchers beyond mediation analysis or causal inference. We share the code for generating these functions as a part of the $\mathsf{DeepMed}$ package.
\end{itemize}

We consider a sample with 10,000 i.i.d. observations. The covariates $X = (X_1, ..., X_p)^{\top}$ are independently drawn from uniform distribution $\mathrm{Uniform} ([-1, 1])$. The outcome $Y$, treatment $D$ and mediator $M$ are generated as follows:
\begin{equation*}
\begin{split}
     D \sim \text{Bernoulli} (\mathfrak{s} (d(X))), M \sim 0.2 D + m (X) + \mathcal{N} (0, 1), Y \sim 0.2 D + M + y (X) + \mathcal{N} (0, 1),
\end{split}
\end{equation*}
where $\mathfrak{s} (x) \coloneqq \left( 1 + e^{-x} \right)^{-1}$, and we consider the following three cases to generate the nonlinear functions $d (x), m(x)$ and $y (x)$ in the main text:

$\bullet$ Case 1 (simple functions):
\begin{align*}
    d (x) = x_1 x_2 + x_3 x_4 x_5 + \sin{x_1}, m (x) = 4 \sum_{i = 1}^5 \sin{3 x_i}, y (x) = (x_1 + x_2)^2 + 5 \sin{\sum_{i = 1}^5 x_i}.
\end{align*}
$\bullet$ Case 2 (composition of simple functions): we simulate more complex interactions among covariates by composing simple functions as follow:
\small
\begin{align*}
    & d (x) = d_2 \circ d_1 \circ d_0 (x_1, \cdots, x_5), \text{ with } d_0 (x_1, \cdots, x_5) = \left( \prod_{i = 1}^2 x_i, \prod_{i = 3}^5 x_i, \prod_{i = 1}^2 \sin{x_i}, \prod_{i = 3}^5 \sin{x_i} \right), \\
    & d_1 (a_1, \cdots, a_4) = \left( \sin(a_1 + a_2), \sin{a_2}, a_3, a_4 \right), \text{and } d_2 (b_1, \cdots, b_4) = 0.5 \sin(b_1 + b_2) + 0.5 (b_3 + b_4), \\
    & m (x) = m_1 \circ m_0 (x_1, \ldots, x_5), \text{with } m_0 (x_1, \cdots, x_5) = \left( \sin{x_1}, \cdots, \sin{x_5} \right), m_1 (a_1, \cdots, a_5) = 5 \sin{\sum_{i = 1}^5 a_i} \\
    & \text{and } y (x) = y_2 \circ y_1 \circ y_0 (x_1, \cdots, x_5),  \text{ with } y_0 (x_1, \cdots, x_5) = \left( \sin{\sum_{i = 1}^2 x_i}, \sin{\sum_{i = 3}^5 x_i}, \sin{\sum_{i = 1}^5 x_i} \right), \\
    & y_1 (a_1, a_2, a_3) = \left( \sin (a_1 + a_2), a_3 \right), \text{and } y_2 (b_1, b_2) = 10 \sin (b_1 + b_2).
\end{align*}
\normalsize
$\bullet$ Case 3 (H\"{o}lder functions): we consider more complex nonlinear functions as follows:
{\small
\begin{align*}
& d (x) = x_1 x_2 + x_3 x_4 x_5 + 0.5 \eta (0.2 x_1; \alpha), m (x) = \sum_{i = 1}^5 \eta \left( 0.5 x_i; \alpha \right), y (x) = x_1 x_2 + 3 \eta \left( 0.2 \sum_{i = 1}^5 x_i; \alpha \right)
\end{align*}}

\noindent where $\eta (x; \alpha) = \sum_{j \in J, l \in \mathbb{Z}} 2^{- j (\alpha + 0.25)} w_{j, l} (x)$ with $J = \{0, 3, 6, 9, 10, 16\}$ and $w_{j, l} (\cdot)$ is the D6 father wavelet functions dilated at resolution $j$ shifted by $l$. By construction, $\eta (x; \alpha) \in \mathcal{H}_{1} (\alpha; B)$ for some known constant $B > 0$ following \citet[Theorem 9.6]{hardle1998wavelets}. Here we set $\alpha = 1.2$ and the intrinsic dimension $k = 1$. Thus we expect the $\mathsf{DeepMed}$ estimators are semiparametric efficient. It is indeed the case based on the columns corresponding to Case 3 in Table \ref{Table 1}, suggesting that DNNs can be adaptive to certain low-dimensional structures.

\begin{remark}
\label{beyond}
 The nuisance functions in Cases 3 -- 5 (see Appendix \ref{app:sim}) are less smooth than what have been considered elsewhere, including \citet{farrell2021deep}, \citet{chen2020causal}, and even \citet{adcock2021gap}, a paper dedicated to exposing the gap between theoretical approximation rates and DNN practice. These nuisance functions are designed to be near the boundary of a H\"{o}lder ball with a given smoothness exponent as we add wavelets at very high resolution in $\eta (x; \alpha)$. This is the assumption under which most of the known statistical properties of DNNs are developed.
\end{remark}

\begin{table*}[htb] 
\caption{The biases, empirical standard errors (SE) and root mean squared errors (RMSE) of the estimated $\tau_{tot}, \tau_{\NDE} (1)$ and $\tau_{\NIE} (1)$, and the coverage probabilities (CP) of their corresponding $95 \%$ confidence intervals. $p = 5$ (no irrelevant covariates). The simulation is based on 200 replicates. The full table including $\hat{\tau}_{\NDE} (0)$ and $\hat{\tau}_{\NIE} (0)$ can be found in Table \ref{Table S1} in the Appendix.}  
\centering
\resizebox{\textwidth}{33mm}{
\begin{tabular}{cccccccccccccccc}
\toprule[ 1pt]
              &        & \multicolumn{4}{c}{Case 1}        &  & \multicolumn{4}{c}{Case 2}       &  & \multicolumn{4}{c}{Case 3}       \\
\cline{3-6} \cline{8-11} \cline{13-16}
           & Method & Bias   & SE    & RMSE  & CP    &  & Bias   & SE    & RMSE  & CP    &  & Bias   & SE    & RMSE  & CP    \\
\midrule[ 1pt]
$\tau_{tot}$     & $\mathsf{DeepMed}$    & -0.001 & 0.032 & 0.032 & 0.945 &  & -0.004 & 0.032 & 0.032 & 0.955 &  & 0.008  & 0.037 & 0.038 & 0.920 \\
                 & Lasso  & 0.192  & 0.089 & 0.212 & 0.460 &  & -0.304 & 0.116 & 0.325 & 0.215 &  & 0.346  & 0.079 & 0.355 & 0.010 \\
                 & RF     & 0.067  & 0.042 & 0.079 & 0.775 &  & -0.078 & 0.056 & 0.096 & 0.950 &  & -0.009 & 0.042 & 0.043 & 0.985 \\
                 & GBM    & -0.015 & 0.036 & 0.039 & 0.940 &  & -0.044 & 0.055 & 0.070 & 0.850 &  & 0.019  & 0.041 & 0.045 & 0.930 \\
                 & Oracle & -0.001 & 0.029 & 0.029 & 0.955 &  & -0.003 & 0.029 & 0.029 & 0.925 &  & -0.001 & 0.032 & 0.032 & 0.935 \\
\hline
$\tau_{\NDE}(1)$ & $\mathsf{DeepMed}$    & 0.000  & 0.027 & 0.027 & 0.945 &  & -0.007 & 0.023 & 0.024 & 0.955 &  & 0.000  & 0.026 & 0.026 & 0.965 \\
                 & Lasso  & 0.130  & 0.043 & 0.137 & 0.220 &  & -0.375 & 0.059 & 0.380 & 0.000 &  & 0.226  & 0.064 & 0.235 & 0.050 \\
                 & RF     & 0.048  & 0.029 & 0.056 & 0.700 &  & -0.188 & 0.044 & 0.193 & 0.005 &  & 0.030  & 0.038 & 0.048 & 0.980 \\
                 & GBM    & -0.040 & 0.031 & 0.051 & 0.770 &  & -0.164 & 0.046 & 0.170 & 0.040 &  & 0.011  & 0.042 & 0.043 & 0.920 \\
                 & Oracle & 0.000  & 0.022 & 0.022 & 0.945 &  & -0.002 & 0.020 & 0.020 & 0.985 &  & 0.001  & 0.022 & 0.022 & 0.955 \\
\hline
$\tau_{\NIE}(1)$ & $\mathsf{DeepMed}$    & -0.001 & 0.025 & 0.025 & 0.960 &  & 0.005  & 0.029 & 0.029 & 0.915 &  & 0.008  & 0.031 & 0.032 & 0.905 \\
                 & Lasso  & 0.058  & 0.077 & 0.096 & 0.875 &  & 0.069  & 0.094 & 0.117 & 0.905 &  & 0.120  & 0.045 & 0.128 & 0.220 \\
                 & RF     & 0.066  & 0.037 & 0.076 & 0.665 &  & 0.108  & 0.059 & 0.123 & 0.860 &  & -0.045 & 0.038 & 0.059 & 0.765 \\
                 & GBM    & 0.023  & 0.031 & 0.039 & 0.890 &  & 0.120  & 0.064 & 0.136 & 0.485 &  & -0.001 & 0.037 & 0.037 & 0.935 \\
                 & Oracle & -0.001 & 0.020 & 0.020 & 0.975 &  & 0.000  & 0.021 & 0.021 & 0.930 &  & -0.002 & 0.022 & 0.022 & 0.920 \\
\bottomrule[ 1pt]
\end{tabular}
}
\label{Table 1}
\end{table*}

In all the above cases, $\tau_{tot} = 0.4$ and $\tau_{\NDE} (d) = \tau_{\NIE}(d) = 0.2$ for $d \in \{0, 1\}$. We also consider the cases where the total number of covariates $p = 20$ and $100$ but only the first five covariates are relevant to $Y$, $M$ and $D$. All simulation results are based on 200 replicates. The sigmoid function is used in the final layer when the response variable is binary. For comparison, we also use the Lasso, random forest (RF) and gradient boosted machine (GBM) to estimate the nuisance functions, and use the true nuisance functions (Oracle) as the benchmark. The Lasso is implemented using the $\mathsf{R}$ package ``hdm'' with a data-driven penalty. The DNN, RF and GBM are implemented using the $\mathsf{R}$ packages ``keras'', ``randomForest'' and ``gbm'', respectively. We adopt a 3-fold cross-validation to choose the hyperparameters for DNNs (depth $L$, width $K$, $L_{1}$-regularization parameter $\lambda$ and epochs), RF (number of trees and maximum number of nodes) and GBM (numbers of trees and depth). We use a completely independent sample for the hyperparameter selection. In this paper, we only use one extra dataset to conduct the cross-validation for hyperparameter selection, so our simulation results are conditional on this extra dataset. We use the cross-entropy loss for the binary response and the mean-squared loss for the continuous response. We fix the batch-size as 100 and the other hyperparameters for the other methods are set to the default values in their $\mathsf{R}$ packages. See Appendix \ref{app:sim} for more details.
   
We compare the performances of different methods in terms of the biases, empirical standard errors (SE) and root mean squared errors (RMSE) of the estimates as well as the coverage probabilities (CP) of their $95 \%$ confidence intervals. When $p = k = 5$ (all covariates are relevant or no low-dimensional structures), $\mathsf{DeepMed}$ has smaller bias and RMSE than the other competing methods, and is only slightly worse than Oracle. Lasso has the largest bias and poor CP as expected since it does not capture the nonlinearity of the nuisance functions. RF and GBM also have substantial biases, especially in Case 2 with compositions of simple functions. Overall, $\mathsf{DeepMed}$ performs better than the competing methods (Table \ref{Table 1}). From the empirical distributions, we can also see that they are nearly unbiased and normally distributed in Cases 1-3 (Figures \ref{fig:case1}-\ref{fig:case3}). When $p = 20$ or $100$ but only the first five covariates are relevant ($k = 5$), $L_{1}$-regularization in the input-layer drastically improves the performance of $\mathsf{DeepMed}$ (Table \ref{Table: no regularization}). $\mathsf{DeepMed}$ with $L_1$-regularization in the input-layer also has smaller bias and RMSE than the other competing methods (Tables \ref{Table:p=20} and \ref{Table:p=100}). 
 
As expected, more precise nuisance function estimates (i.e., smaller validation loss) generally lead to more precise causal effect estimates. The validation losses of nuisance function estimates from $\mathsf{DeepMed}$ are generally much smaller than those using Lasso, RF and GBM (Tables \ref{Table: cv loss/p=5}-\ref{Table: cv loss/p=100}). 
 

\begin{remark}
\label{fp}
Due to space limitations, we defer Cases 4, 5 to Appendix \ref{app:sim}, in which $\mathsf{DeepMed}$ fails to be semiparametric efficient, compared to the Oracle; see an extended discussion in Appendix \ref{app:sim}. We conjecture this may be due to the implicit regularization of gradient-based training algorithm such as SGD (Table \ref{Table:SGD}) or $\mathsf{adam}$ \citep{kingma2015adam} (all simulation results except Table \ref{Table:SGD}), which is used to train the DNNs to estimate the nuisance parameters, instead of actually solving the ERM \eqref{erm}. 
Most previous works focus on the benefit of implicit regularization \citep{neyshabur2017implicit, bartlett2020benign} on generalization. Yet, implicit regularization might inject implicit bias into causal effect estimates, which could make statistical inference invalid. Such a potential curse of implicit regularization has not been documented in the DNN-based causal inference literature before and exemplify the value of our synthetic experiments. We believe this is an important open research direction for theoretical results to better capture the empirical performance of DNN-based causal inference methods such as $\mathsf{DeepMed}$.
\end{remark}

\section{Real data analysis on fairness}
\label{sec:real}

As a proof of concept, we use $\mathsf{DeepMed}$ and other competing methods to re-analyze the COMPAS algorithm \citep{dressel2018accuracy}. In particular, we are interested in the NDE of race $D$ on the recidivism risk (or the COMPAS score) $Y$ with the number of prior convictions as the mediator $M$. For race, we mainly focus on the Caucasians population ($D = 0$) and the African-Americans population ($D = 1$), and exclude the individuals of other ethnicity groups. The COMPAS score ($Y$) is ordinal, ranging from 1 to 10 (1: lowest risk; 10: highest risk). We also include the demographic information (age and gender) as covariates $X$.
 
All the methods find significant positive NDE of race on the COMPAS score at $\alpha$-level 0.005 (Table \ref{Table real2}; all p-values $< 10^{-7}$), consistent with previous findings \citep{nabi2018fair}. Thus the COMPAS algorithm tends to assign higher recidivism risks to African-Americans than to Caucasians, even when they have the same number of prior convictions. 
The validation losses of nuisance function estimates by $\mathsf{DeepMed}$ are smaller than the other competing methods (Table \ref{Table real2:validation loss}), possibly suggesting smaller biases of the corresponding NDE/NIE estimators.

We emphasize that research in machine learning fairness should be held accountable \citep{bao2021s}. Our data analysis is merely a proof-of-concept that $\mathsf{DeepMed}$ works in practice and the conclusion from our data analysis should not be treated as definitive. We defer the comments on potential issues of unmeasured confounding to Appendix \ref{app:endogeneity} and another real data analysis to Appendix \ref{app:real}.

\begin{table}[h]
\caption{Results for real data application to COMPAS algorithm fairness.}  
\centering
{\small\begin{tabular}{cccc|cccccccccccc}
\toprule[ 1pt]
  
          Method & Effect & Estimate   & SE & Method & Effect & Estimate   & SE \\
\midrule[ 1pt]
    & $\tau_{tot}$    & 1.136 & 0.069 & & $\tau_{tot}$    & 1.083 & 0.111 \\ 
      & $\tau_{\NDE}(1)$ & 0.564 & 0.068 & & $\tau_{\NDE}(1)$ & 0.589 & 0.070 \\ 
      $\mathsf{DeepMed}$ & $\tau_{\NDE}(0)$ & 0.524 & 0.062 & RF & $\tau_{\NDE}(0)$ & 0.569 & 0.103 \\ 
      & $\tau_{\NIE}(1)$ & 0.612 & 0.042 & & $\tau_{\NIE}(1)$ & 0.514 & 0.049 \\ 
      & $\tau_{\NIE}(0)$ & 0.572 & 0.051 & & $\tau_{\NIE}(0)$ & 0.494 & 0.065 \\ 
      \hline
      & $\tau_{tot}$    & 1.150 & 0.068 & & $\tau_{tot}$    & 1.180 & 0.068 \\ 
      & $\tau_{\NDE}(1)$ & 0.575 & 0.063 & & $\tau_{\NDE}(1)$ & 0.550 & 0.063 \\ 
      Lasso & $\tau_{\NDE}(0)$ & 0.587 & 0.062 & GBM & $\tau_{\NDE}(0)$ & 0.526 & 0.061 \\ 
      & $\tau_{\NIE}(1)$ & 0.563 & 0.032 & & $\tau_{\NIE}(1)$ & 0.654 & 0.041 \\ 
      & $\tau_{\NIE}(0)$ & 0.575 & 0.040 & & $\tau_{\NIE}(0)$ & 0.630 & 0.044 \\ 
\bottomrule[ 1pt]
\end{tabular}} \label{Table real2}
\end{table}
\section{Conclusion and Discussion}
\label{sec:disc}

In this paper, we proposed $\mathsf{DeepMed}$ for semiparametric mediation analysis with DNNs. We established novel statistical properties for DNN-based causal effect estimation that can (1) circumvent sparse DNN architectures and (2) leverage certain low-dimensional structures of the nuisance functions. These results significantly advance our current understanding of DNN-based causal inference including mediation analysis.

Evaluated by our extensive synthetic experiments, $\mathsf{DeepMed}$ mostly exhibits improved finite-sample performance over the other competing machine learning methods. But as mentioned in Remark \ref{fp}, there is still a large gap between statistical guarantees and empirical observations. Therefore an important future direction is to incorporate the training process while investigating the statistical properties to have a deeper theoretical understanding of DNN-based causal inference. It is also of future research interests to enable $\mathsf{DeepMed}$ to handle unmeasured confounding and more complex path-specific effects \citep{malinsky2019potential, miles2020semiparametric}, and incorporate other hyperparameter tuning strategies that leverage the multiply-robustness property, such as the minimax criterion \citep{robins2020double, cui2019selective}. 

Finally, we warn readers that all causal inference methods, including $\mathsf{DeepMed}$, may have negative societal impact if they are used without carefully checking their working assumptions.



\section*{Acknowledgement and Disclosure of Funding}
The authors thank four anonymous reviewers and one anonymous area chair for helpful comments, \href{https://gaofn.xyz/}{Fengnan Gao} for some initial discussion on how to incorporate low-dimensional manifold assumptions using DNNs and \href{https://scholar.google.com/citations?user=Ys5ZVhEAAAAJ&hl=en}{Ling Guo} for discussion on DNN training. The authors would also like to thank Department of Statistics and Actuarial Sciences at The University of Hong Kong for providing high-performance computing servers that supported the numerical experiments in this paper. L. Liu gratefully acknowledges funding support by Natural Science Foundation of China Grant No.12101397 and No.12090024, Pujiang National Lab Grant No. P22KN00524, Natural Science Foundation of Shanghai Grant No.21ZR1431000, Shanghai Science and Technology Commission Grant No.21JC1402900, Shanghai Municipal Science and Technology Major Project No.2021SHZDZX0102, and Shanghai Pujiang Program Research Grant No.20PJ140890.

\bibliographystyle{plainnat}
\bibliography{Master.bib}

\newpage

\appendix
\onecolumn
\setcounter{table}{0}
\setcounter{figure}{0}
\renewcommand{\thetable}{A\arabic{table}}
\renewcommand{\thefigure}{A\arabic{figure}}
\newgeometry{margin=1in}



\newpage
\textbf{\Large Appendix}
\section{More comments on the Ignorability conditions}\label{app:ignore}

It is well known that NDE/NIE is not nonparametrically identifiable without assuming the four ignorability conditions listed in Assumption ii: for all $d, d' \in \{0, 1\}$ and $m \in \mathcal{M}$
\begin{align*}
    & \text{No unmeasured treatment-outcome confounding}: Y (d, m) \perp D | X; \\
    & \text{No unmeasured treatment-mediator confounding}: Y (d, m) \perp M | X, D; \\
    & \text{No unmeasured treatment-mediator confounding}: M (d) \perp D | X; \\
    & \text{Cross-world condition}: Y (d, m) \perp M (d') | X.
\end{align*}

The first three are standard ignorability conditions; but the fourth one involves ``cross-world'' potential outcomes $Y (d, m)$ and $M (d')$ when $d \neq d'$. The cross-world assumption is often criticized by researchers who are ``interventionists'' \citepSM{robins2022interventionist} because this condition cannot be empirically verified even by conducting randomized trials. To resolve this issue, many other direct/indirect effects are developed that are identifiable without assuming the cross-world condition, e.g. the interventional direct/indirect effect (IDE/IIE) \citepSM{vanderweele2014effect}. We decided to focus on the more standard NDE/NIE in this paper because the identification formulae of NDE and NIE as in \eqref{eq:id} are the same as those of IDE and IIE. It is beyond the scope of this paper to discuss the conceptual (dis)advantages of different types of direct/indirect effects for mediation analysis.

\section{The bias of generic sample-splitting multiply-robust estimators of NDE/NIE}\label{app:bias}
In this section, we prove Proposition \ref{thm:master}, which is a consequence of the Proposition below.
\begin{proposition}\label{prop:mr}
Conditional on the nuisance sample data $\mathcal{D}_{\nu}$, the bias of $\tilde{\phi} (d, d')$ as an estimator of $\phi (d, d')$ is of the following second-order form:
\begin{equation*}
\begin{split}
& \E \left[ \tilde{\phi} (d, d') - \phi (d, d') | \mathcal{D}_{\nu} \right] \\
& = \E_{X} \left[ \underset{m \in \mathcal{M}}{\int} \left( 1 - \frac{a (d' | X)}{\tilde{a} (d' | X)} \right) \left( \frac{\tilde{f} (m | X, d')}{f (m | X, d')} - 1 \right) \tilde{\mu} (X, d, m) f (m | X, d') \diff m \right] \\
& + \E_{X} \left[ \underset{m \in \mathcal{M}}{\int} \left( 1 - \frac{f (m | X, d)}{\tilde{f} (m | X, d)} \frac{\tilde{f} (m | X, d')}{f (m | X, d')} \right) \left( \tilde{\mu} (X, d, m) - \mu (X, d, m) \right) f (m | X, d') \diff m \right] \\
& + \E_{X} \left[ \underset{m \in \mathcal{M}}{\int} \left( 1 - \frac{a (d | X)}{\tilde{a} (d | X)} \right) \frac{\tilde{f} (m | X, d')}{\tilde{f} (m | X, d)} \left( \tilde{\mu} (X, d, m) - \mu (X, d, m) \right) f (m | X, d) \diff m \right].
\end{split}
\end{equation*}
Consequently, one obtains the following upper bound of the bias:
\begin{equation}\label{upper}
\begin{split}
& \mathrm{Bias} (\tilde{\phi} (d, d')) \equiv \left\vert \E \left[ \tilde{\phi} (d, d') - \phi (d, d') | \mathcal{D}_{\nu} \right] \right\vert \\
\lesssim & \ r_{a, d} \cdot r_{f, d} + \max_{d'' \in \{0, 1\}} r_{f, d''} \cdot r_{\mu, d} + r_{a, d} \cdot r_{\mu, d}.
\end{split}
\end{equation}
\end{proposition}

\begin{proof}
The first statement on the bias follows directly from sample-splitting and the form of the EIF $\psi_{d, d'} (o) - \phi (d, d')$. The second statement is obtained by the application of triangle inequality and Cauchy-Schwarz inequality.
\end{proof}

It is worth noting that the upper bound in \eqref{upper} by Cauchy-Schwarz inequality is by no means the only analysis strategy. For instance, one could also upper bound the bias by H\"{o}lder inequality if convergence rates of DNN-based nuisance function estimators are available in general $L_{p}$-norms beyond $p = 2$. Proposition \ref{thm:master} is a generalization of the results in \citetSM{robins2008higher, chernozhukov2018double} to mediation analysis.

\section{H\"{o}lder functions, their ERM DNN-based estimators and statistical properties}
\label{app:holder}
As in the main text, we denote $\mathcal{H}_{p} (\alpha; C)$ as the H\"{o}lder balls of functions from $\mathbb{R}^{p}$ to $\mathbb{R}$, with smoothness exponent $\alpha$ and radii $C$ \citepSM{triebel2010theory, gine2016mathematical}, formally defined below:
\begin{equation*}
\begin{split}
\mathcal{H}_{p} (\alpha; C) \coloneqq \left\{ \begin{array}{ll}
\left\{ g: [0, 1]^{p} \rightarrow \mathbb{R}; \left. \begin{array}{c}
\underset{m \in \mathbb{Z}_{\geq 0}^{p}, |m|_{1} < \lfloor \alpha \rfloor}{\max} \Vert \partial^{m} g \Vert_{\infty} \leq C \\
\text{ and } \underset{m \in \mathbb{Z}_{\geq 0}^{p}, |m|_{1} = \lfloor \alpha \rfloor}{\max} \underset{w, w' \in [0, 1]^{p}, w \neq w'}{\sup} \dfrac{\vert \partial^{m} g (w) - \partial^{m} g (w') \vert}{\Vert w - w' \Vert_{\infty}^{\alpha - \lfloor \alpha \rfloor}} \leq C
\end{array} \right. \right\} & \alpha \geq 1, \\
\left\{ g: [0, 1]^{p} \rightarrow \mathbb{R}; \underset{w, w' \in [0, 1]^{p}, w \neq w'}{\sup} \dfrac{\vert g (w) - g (w') \vert}{\Vert w - w' \Vert_{\infty}^{\alpha}} \leq C \right\} & 0 < \alpha < 1.
\end{array} \right.
\end{split}
\end{equation*}
It is well-known \citepSM{stone1982optimal} that the minimax optimal convergence rate of estimating $g \in \mathcal{H}_{p} (\alpha; C)$ in $L_{2}$-norm is $n^{- \frac{\alpha}{2 \alpha + p}}$, suffering from curse-of-dimensionality. As mentioned in the main text, one possibility is to consider the function space $\calH_{k}^{\dag} (\alpha; C)$ by assuming that the nuisance functions only depend on the covariates $w \in \mathbb{R}^{p}$ via a $k$-dimensional linear subspace $\Gamma w$, where $\Gamma \in \mathbb{R}^{k \times p}$ and is unknown.

There exist many estimators attaining the optimal rate: e.g. wavelet projection estimators, kernel estimators, etc. In particular, sparse DNN-based estimators have also shown to attain the optimal rate up to a log-factor \citepSM{schmidt2020nonparametric, suzuki2019adaptivity}. However, since sparse DNNs are computationally demanding to search over $\calF_{nn}$, we prefer results that avoid such sparsity constraints. To this end, it is easy to show the following by adapting the proof of Theorem 1.1 of \citetSM{lu2021deep}:
\begin{lemma}\label{holder_approx}
Given $g \in \calH_{k}^{\dag} (\alpha; C)$, for large enough depth and width $L, K \in \mathbb{Z}_{> 0}$ and some known constant $B > 0$, there exists $\tilde{g} \in \calF_{nn} (L, K, B)$ such that
$$
\Vert g - \tilde{g} \Vert_{\infty} \lesssim \left( \frac{L}{\log L} \frac{K}{\log K} \right)^{- 2 \alpha / k}.
$$
\end{lemma}

The proof is straightforward by simply taking the parameter of the input layer to be $W^{(1)} = (\Gamma, -\Gamma) \in \mathbb{R}^{2 k \times p}$ and $b^{(1)} = 0$ and the second layer parameters chosen appropriately such that the input becomes $\Gamma x$ before the ReLU activation function. The rest of the proof then follows directly by applying Theorem 1.1 of \citetSM{lu2021deep}.

Next, we invoke the metric entropy bound of $\calF_{nn} (L, K, B)$ established by Lemma 3 of \citetSM{suzuki2019adaptivity}:
\begin{lemma}[Metric entropy bound of $\calF_{nn} (L, K, B)$]\label{me}
Denote the covering number \citepSM{van1996weak} of $\calF_{nn} (L, K, B)$ w.r.t. $L_{\infty}$-norm as $N (\epsilon, \calF_{nn} (L, K, B), \Vert \cdot \Vert_{\infty})$. Then for any $\epsilon > 0$, for large enough $L, K \in \mathbb{Z}_{> 0}$ and $B > 0$, we have
\begin{align*}
    \log N (\epsilon, \calF_{nn} (L, K, B), \Vert \cdot \Vert_{\infty}) \lesssim (L K)^{2} \log \left( \frac{L K}{\epsilon} \right).
\end{align*}
\end{lemma}

Combining the above two lemmas, we are now ready to prove Lemma \ref{lem:holder}.

\begin{proof}[Proof of Lemma \ref{lem:holder}]
Following Lemma 3.2 of \citetSM{jiao2021deep} or standard $M$-estimation and empirical process theory \citepSM{van1996weak}, under sub-Gaussian assumption of the noise $\xi$, for the ERM estimator $\hat{g}$ given in \eqref{erm}, we have
\begin{align*}
    \sup_{g \in \calH_{k}^{\dag} (\alpha; C)} \E \left[ (\hat{g} (W) - g (W))^{2} \right] & \lesssim \frac{(\log n)^{2} \log N (1 / n, \calF_{nn} (L, K, B), \Vert \cdot \Vert_{\infty})}{n} + \inf_{\tilde{g} \in \calF_{nn} (L, K, B)} \Vert \tilde{g} - g \Vert_{\infty}^{2} \\
    & \lesssim \frac{(\log n)^{3} (L K)^{2} \log (L K)}{n} + \left( \frac{L}{\log L} \frac{K}{\log K} \right)^{- 4 \alpha / k},
\end{align*}
where the second inequality follows from Lemma \ref{holder_approx} and \ref{me}.

Finally, with a simple bias-variance trade-off argument, we can choose $L K \asymp n^{\frac{k}{2 (k + 2 \alpha)}}$ to obtain the desired rate.
\end{proof}

Before proceeding, we make the following remark regarding the optimality of the results in Theorem \ref{thm:holder}.
\begin{remark}
We believe the conditions in Theorem \ref{thm:holder} are not tight. For example, when the nuisance functions belong to certain H\"{o}lder balls, the sufficient and necessary H\"{o}lder-type condition for the existence of semiparametric efficient estimator of $\tau_{tot}$ is $\frac{\alpha_{a} + \alpha_{\mu}}{2 k} > \frac{1}{4}$. The infinite-order $U$-statistic estimator of \citetSM{mukherjee2017semiparametric} is the only known semiparametric efficient estimator under the above minimal H\"{o}lder-type condition. Yet, their estimators require delicate regularity properties of the estimated nuisance functions, which are difficult to verify for DNNs. It is an interesting open theoretical problem how to achieve semiparametric efficiency under minimal H\"{o}lder-type condition even simply for $\tau_{tot}$, when the nuisance functions are estimated by DNNs.
\end{remark}

It is possible to generalize H\"{o}lder balls in several directions: e.g. assuming different smoothness exponents in different dimensions of the input \citepSM{suzuki2019adaptivity} or composing H\"{o}lder functions hierarchically to mimick the composition structure of DNNs \citepSM{schmidt2020nonparametric} (e.g. Case 5 in Appendix \ref{app:sim}).

\begin{proof}[Proof of Theorem \ref{thm:holder}]
Theorem \ref{thm:holder} is an immediate consequence of Lemma \ref{lem:holder} and Proposition \ref{thm:master}.
\end{proof}

\section{Tables and figures related to the main text}
\label{supp_tab_fig}

\begin{figure*}[hbt!]
\centering
\includegraphics[scale=0.8]{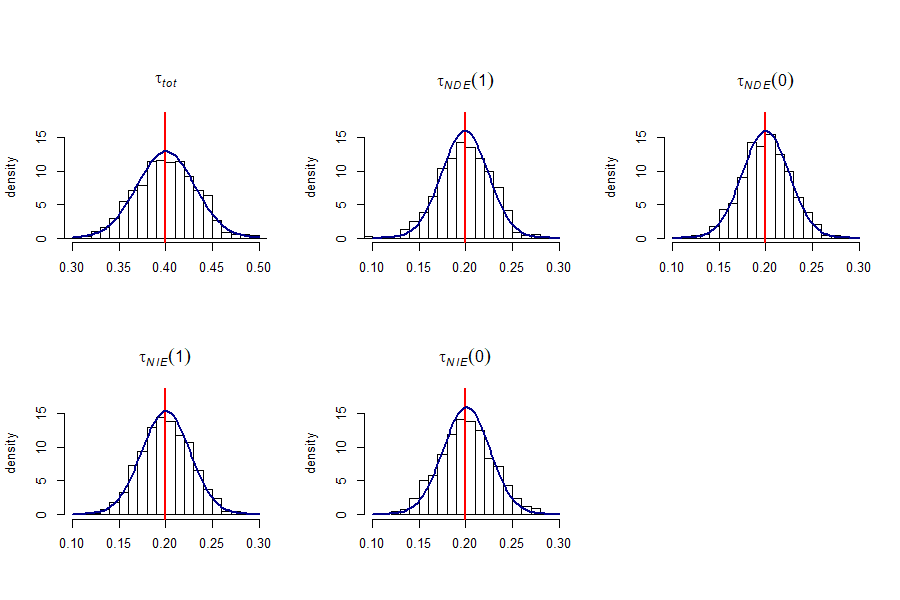}
\caption{(Case 1) The empirical distributions of the estimated total effects, NIE and NDE by $\mathsf{DeepMed}$. The results are based on 1000 simulation replicates and the number of covariates $p = 5$. The red vertical lines indicate the true effects. The blue curves represent the normal density with the means at the true effects and the estimated standard errors.}
\label{fig:case1}
\end{figure*}

\begin{figure*}[hbt!]
\centering
\includegraphics[scale=0.7]{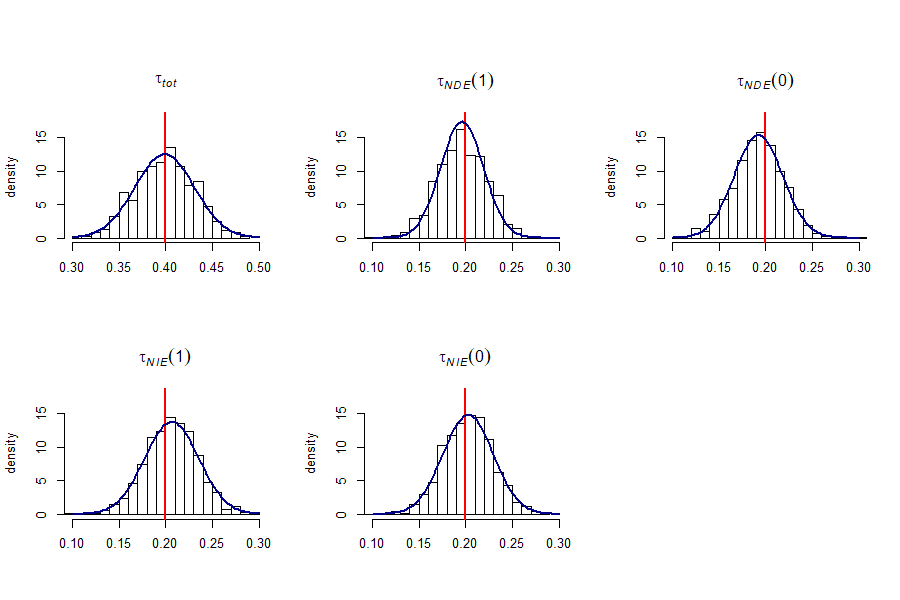}
\caption{(Case 2) The empirical distributions of the estimated total effects, NIE and NDE by $\mathsf{DeepMed}$. The results are based on 1000 simulation replicates and the number of covariates $p = 5$. The red vertical lines indicate the true effects. The blue curves represent the normal density with the means at the true effects and the estimated standard errors.}
\label{fig:case2}
\end{figure*}

\begin{figure*}[hbt!]
\centering
\includegraphics[scale=0.7]{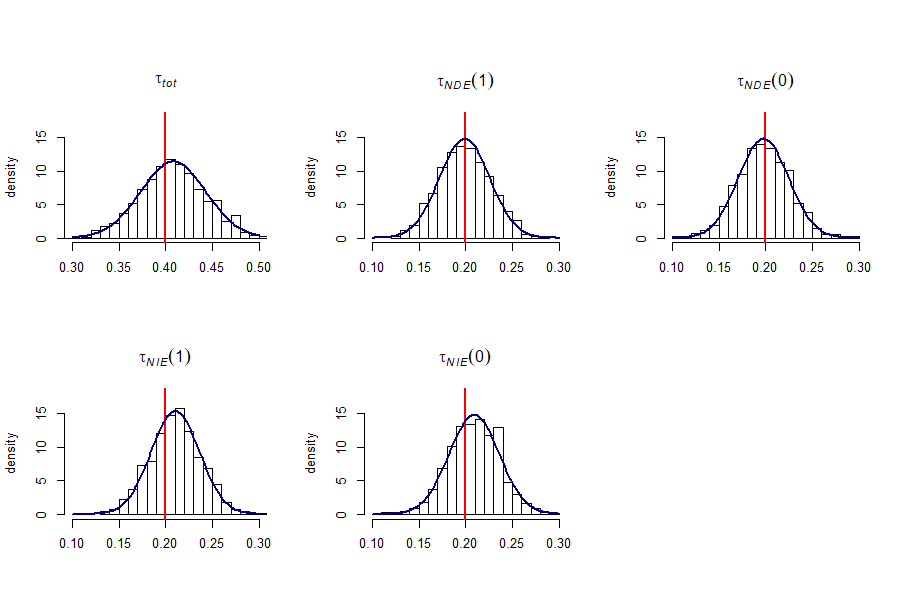}
\caption{(Case 3) The empirical distributions of the estimated total effects, NIE and NDE by $\mathsf{DeepMed}$. The results are based on 1000 simulation replicates and the number of covariates $p = 5$. The red vertical lines indicate the true effects. The blue curves represent the normal density with the means at the true effects and the estimated standard errors.}
\label{fig:case3}
\end{figure*}

In this section, we collect tables and figures that are related to Cases 1 -- 3 of the simulated experiments and real data analysis of the COMPAS dataset, including Table \ref{Table S1} to Table \ref{Table real2:validation loss} and Figure \ref{fig:case1} to Figure \ref{fig:case3}.

\begin{table*}[hbt!] 
\caption{The biases, empirical standard errors (SE) and root mean squared errors (RMSE) of the estimated average total effects, NDE, and NIE, and the coverage probabilities (CP) of their corresponding $95 \%$ confidence intervals. $p = 5$ (no irrelevant covariates). The simulation is based on 200 replicates.}  
\centering
\resizebox{\textwidth}{60mm}{
\begin{tabular}{cccccccccccccccc}
\toprule[ 1pt]
              &        & \multicolumn{4}{c}{Case 1}        &  & \multicolumn{4}{c}{Case 2}       &  & \multicolumn{4}{c}{Case 3}       \\
\cline{3-6} \cline{8-11} \cline{13-16}
           & Method & Bias   & SE    & RMSE  & CP    &  & Bias   & SE    & RMSE  & CP    &  & Bias   & SE    & RMSE  & CP    \\
\midrule[ 1pt]
$\tau_{tot}$     & $\mathsf{DeepMed}$    & -0.001 & 0.032 & 0.032 & 0.945 &  & -0.004 & 0.032 & 0.032 & 0.955 &  & 0.008  & 0.037 & 0.038 & 0.920 \\
                 & Lasso  & 0.192  & 0.089 & 0.212 & 0.460 &  & -0.304 & 0.116 & 0.325 & 0.215 &  & 0.346  & 0.079 & 0.355 & 0.010 \\
                 & RF     & 0.067  & 0.042 & 0.079 & 0.775 &  & -0.078 & 0.056 & 0.096 & 0.950 &  & -0.009 & 0.042 & 0.043 & 0.985 \\
                 & GBM    & -0.015 & 0.036 & 0.039 & 0.940 &  & -0.044 & 0.055 & 0.070 & 0.850 &  & 0.019  & 0.041 & 0.045 & 0.930 \\
                 & Oracle & -0.001 & 0.029 & 0.029 & 0.955 &  & -0.003 & 0.029 & 0.029 & 0.925 &  & -0.001 & 0.032 & 0.032 & 0.935 \\
\hline
$\tau_{\NDE}(1)$ & $\mathsf{DeepMed}$    & 0.000  & 0.027 & 0.027 & 0.945 &  & -0.007 & 0.023 & 0.024 & 0.955 &  & 0.000  & 0.026 & 0.026 & 0.965 \\
                 & Lasso  & 0.130  & 0.043 & 0.137 & 0.220 &  & -0.375 & 0.059 & 0.380 & 0.000 &  & 0.226  & 0.064 & 0.235 & 0.050 \\
                 & RF     & 0.048  & 0.029 & 0.056 & 0.700 &  & -0.188 & 0.044 & 0.193 & 0.005 &  & 0.030  & 0.038 & 0.048 & 0.980 \\
                 & GBM    & -0.040 & 0.031 & 0.051 & 0.770 &  & -0.164 & 0.046 & 0.170 & 0.040 &  & 0.011  & 0.042 & 0.043 & 0.920 \\
                 & Oracle & 0.000  & 0.022 & 0.022 & 0.945 &  & -0.002 & 0.020 & 0.020 & 0.985 &  & 0.001  & 0.022 & 0.022 & 0.955 \\
\hline
$\tau_{\NDE}(0)$ & $\mathsf{DeepMed}$    & 0.000  & 0.025 & 0.025 & 0.940 &  & -0.009 & 0.026 & 0.028 & 0.915 &  & -0.001 & 0.027 & 0.027 & 0.940 \\
                 & Lasso  & 0.134  & 0.043 & 0.141 & 0.170 &  & -0.373 & 0.058 & 0.377 & 0.000 &  & 0.227  & 0.064 & 0.236 & 0.045 \\
                 & RF     & 0.001  & 0.030 & 0.030 & 0.970 &  & -0.186 & 0.047 & 0.192 & 0.020 &  & 0.036  & 0.036 & 0.051 & 0.950 \\
                 & GBM    & -0.037 & 0.030 & 0.048 & 0.800 &  & -0.164 & 0.049 & 0.171 & 0.055 &  & 0.020  & 0.044 & 0.048 & 0.920 \\
                 & Oracle & 0.000  & 0.022 & 0.022 & 0.955 &  & -0.002 & 0.019 & 0.019 & 0.985 &  & 0.001  & 0.022 & 0.022 & 0.950 \\
\hline
$\tau_{\NIE}(1)$ & $\mathsf{DeepMed}$    & -0.001 & 0.025 & 0.025 & 0.960 &  & 0.005  & 0.029 & 0.029 & 0.915 &  & 0.008  & 0.031 & 0.032 & 0.905 \\
                 & Lasso  & 0.058  & 0.077 & 0.096 & 0.875 &  & 0.069  & 0.094 & 0.117 & 0.905 &  & 0.120  & 0.045 & 0.128 & 0.220 \\
                 & RF     & 0.066  & 0.037 & 0.076 & 0.665 &  & 0.108  & 0.059 & 0.123 & 0.860 &  & -0.045 & 0.038 & 0.059 & 0.765 \\
                 & GBM    & 0.023  & 0.031 & 0.039 & 0.890 &  & 0.120  & 0.064 & 0.136 & 0.485 &  & -0.001 & 0.037 & 0.037 & 0.935 \\
                 & Oracle & -0.001 & 0.020 & 0.020 & 0.975 &  & 0.000  & 0.021 & 0.021 & 0.930 &  & -0.002 & 0.022 & 0.022 & 0.920 \\
\hline
$\tau_{\NIE}(0)$ & $\mathsf{DeepMed}$    & -0.001 & 0.028 & 0.028 & 0.940 &  & 0.003  & 0.027 & 0.027 & 0.930 &  & 0.008  & 0.029 & 0.030 & 0.910 \\
                 & Lasso  & 0.062  & 0.078 & 0.100 & 0.870 &  & 0.071  & 0.095 & 0.119 & 0.905 &  & 0.120  & 0.045 & 0.128 & 0.220 \\
                 & RF     & 0.019  & 0.037 & 0.042 & 0.935 &  & 0.110  & 0.058 & 0.124 & 0.835 &  & -0.038 & 0.036 & 0.052 & 0.845 \\
                 & GBM    & 0.025  & 0.031 & 0.040 & 0.910 &  & 0.120  & 0.059 & 0.134 & 0.495 &  & 0.008  & 0.037 & 0.038 & 0.935 \\
                 & Oracle & -0.001 & 0.019 & 0.019 & 0.980 &  & 0.000  & 0.022 & 0.022 & 0.920 &  & -0.001 & 0.023 & 0.023 & 0.930\\
\bottomrule[ 1pt]
\end{tabular}
}
\label{Table S1}
\end{table*}

\begin{table*}[p]
\caption{The biases, empirical standard errors (SE) and root mean squared errors (RMSE) of the total effects, NDE and NIE, and the coverage probabilities (CP) of their corresponding $95 \%$ confidence intervals. There exist irrelevant covariates in this setup ($p = 100$). The simulation is based on 200 replicates.}  
\centering
\begin{tabular}{m{2.5cm}<{\centering}m{1.3cm}<{\centering}m{1cm}<{\centering}m{1cm}<{\centering}m{1cm}<{\centering}m{1cm}<{\centering}m{0.1cm}<{\centering}m{1cm}<{\centering}m{1cm}<{\centering}m{1cm}<{\centering}m{1cm}<{\centering}}
\toprule[ 1pt]
   &  & \multicolumn{4}{c}{Input-layer $L_{1}$ regularization}     &  & \multicolumn{4}{c}{No regularization}     \\
   \cline{3-6} \cline{8-11}
   &      & Bias   & SE    & RMSE  & CP    &  & Bias   & SE    & RMSE  & CP    \\
\midrule[ 1pt]     
Case 1 & $\tau_{tot}$     & 0.008  & 0.046 & 0.047 & 0.955 &  & 0.155  & 0.086 & 0.177 & 0.540 \\
       & $\tau_{\NDE}(1)$ & 0.009  & 0.043 & 0.044 & 0.920 &  & 0.061  & 0.044 & 0.075 & 0.745 \\
       & $\tau_{\NDE}(0)$ & 0.006  & 0.040 & 0.040 & 0.920 &  & 0.002  & 0.049 & 0.049 & 0.920 \\
       & $\tau_{\NIE}(1)$ & 0.002  & 0.050 & 0.050 & 0.930 &  & 0.152  & 0.090 & 0.177 & 0.550 \\
       & $\tau_{\NIE}(0)$ & -0.001 & 0.044 & 0.044 & 0.945 &  & 0.094  & 0.081 & 0.124 & 0.800 \\
\hline
Case 2 & $\tau_{tot}$     & -0.018 & 0.033 & 0.038 & 0.895 &  & -0.033 & 0.035 & 0.048 & 0.875 \\
       & $\tau_{\NDE}(1)$ & -0.025 & 0.036 & 0.044 & 0.855 &  & -0.266 & 0.045 & 0.270 & 0.000 \\
       & $\tau_{\NDE}(0)$ & -0.018 & 0.036 & 0.040 & 0.855 &  & -0.306 & 0.051 & 0.310 & 0.000 \\
       & $\tau_{\NIE}(1)$ & 0.000  & 0.036 & 0.036 & 0.950 &  & 0.273  & 0.060 & 0.280 & 0.000 \\
       & $\tau_{\NIE}(0)$ & 0.007  & 0.037 & 0.038 & 0.920 &  & 0.233  & 0.055 & 0.239 & 0.005 \\
\hline
Case 3 & $\tau_{tot}$     & 0.019  & 0.043 & 0.047 & 0.915 &  & 0.224  & 0.055 & 0.231 & 0.015 \\
       & $\tau_{\NDE}(1)$ & 0.013  & 0.035 & 0.037 & 0.920 &  & 0.075  & 0.051 & 0.091 & 0.600 \\
       & $\tau_{\NDE}(0)$ & 0.016  & 0.033 & 0.037 & 0.925 &  & 0.089  & 0.054 & 0.104 & 0.515 \\
       & $\tau_{\NIE}(1)$ & 0.003  & 0.038 & 0.038 & 0.940 &  & 0.135  & 0.064 & 0.149 & 0.340 \\
       & $\tau_{\NIE}(0)$ & 0.006  & 0.036 & 0.036 & 0.960 &  & 0.149  & 0.051 & 0.157 & 0.155\\
\bottomrule[ 1pt]
\end{tabular}

\label{Table: no regularization}
\end{table*}


\begin{table*}[hbt!]
\caption{The biases, empirical standard errors (SE) and root mean squared errors (RMSE) of the estimated total effects, NDE, and NIE, and the coverage probabilities (CP) of their corresponding $95 \%$ confidence intervals. There exist irrelevant covariates in this setup ($p = 20$). The simulation is based on 200 replicates.}  
\centering
 \resizebox{\textwidth}{60mm}{
\begin{tabular}{cccccccccccccccc}
\toprule[ 1pt]
              &        & \multicolumn{4}{c}{Case 1}        &  & \multicolumn{4}{c}{Case 2}       &  & \multicolumn{4}{c}{Case 3}       \\
\cline{3-6} \cline{8-11} \cline{13-16}
           & Method & Bias   & SE    & RMSE  & CP    &  & Bias   & SE    & RMSE  & CP    &  & Bias   & SE    & RMSE  & CP    \\
\midrule[ 1pt]
$\tau_{tot}$     & $\mathsf{DeepMed}$    & 0.001  & 0.041 & 0.041 & 0.950 &  & -0.019 & 0.030 & 0.036 & 0.920 &  & 0.010  & 0.036 & 0.037 & 0.945 \\
                 & Lasso  & 0.191  & 0.091 & 0.212 & 0.510 &  & -0.318 & 0.108 & 0.336 & 0.155 &  & 0.334  & 0.077 & 0.343 & 0.010 \\
                 & RF     & 0.035  & 0.049 & 0.060 & 0.955 &  & -0.187 & 0.076 & 0.202 & 0.475 &  & 0.100  & 0.052 & 0.113 & 0.545 \\
                 & GBM    & -0.020 & 0.039 & 0.044 & 0.910 &  & -0.139 & 0.063 & 0.153 & 0.445 &  & 0.050  & 0.050 & 0.071 & 0.800 \\
                 & Oracle & -0.002 & 0.032 & 0.032 & 0.925 &  & -0.003 & 0.029 & 0.029 & 0.925 &  & -0.003 & 0.030 & 0.030 & 0.950 \\
\hline
$\tau_{\NDE}(1)$ & $\mathsf{DeepMed}$    & 0.001  & 0.033 & 0.033 & 0.915 &  & -0.021 & 0.026 & 0.033 & 0.875 &  & 0.003  & 0.028 & 0.028 & 0.955 \\
                 & Lasso  & 0.129  & 0.048 & 0.138 & 0.205 &  & -0.378 & 0.060 & 0.383 & 0.000 &  & 0.216  & 0.062 & 0.225 & 0.065 \\
                 & RF     & 0.033  & 0.037 & 0.050 & 0.890 &  & -0.213 & 0.043 & 0.217 & 0.000 &  & 0.088  & 0.047 & 0.100 & 0.545 \\
                 & GBM    & -0.054 & 0.036 & 0.065 & 0.675 &  & -0.228 & 0.049 & 0.233 & 0.005 &  & 0.038  & 0.049 & 0.062 & 0.905 \\
                 & Oracle & -0.002 & 0.023 & 0.023 & 0.925 &  & -0.002 & 0.020 & 0.020 & 0.985 &  & -0.002 & 0.020 & 0.020 & 0.970 \\
\hline
$\tau_{\NDE}(0)$ & $\mathsf{DeepMed}$    & 0.010  & 0.037 & 0.038 & 0.900 &  & -0.019 & 0.028 & 0.034 & 0.890 &  & -0.001 & 0.029 & 0.029 & 0.945 \\
                 & Lasso  & 0.133  & 0.048 & 0.141 & 0.185 &  & -0.376 & 0.060 & 0.381 & 0.000 &  & 0.216  & 0.064 & 0.225 & 0.075 \\
                 & RF     & 0.007  & 0.038 & 0.039 & 0.955 &  & -0.212 & 0.044 & 0.217 & 0.000 &  & 0.081  & 0.046 & 0.093 & 0.605 \\
                 & GBM    & -0.054 & 0.038 & 0.066 & 0.675 &  & -0.233 & 0.051 & 0.239 & 0.000 &  & 0.044  & 0.051 & 0.067 & 0.860 \\
                 & Oracle & -0.002 & 0.023 & 0.023 & 0.930 &  & -0.002 & 0.019 & 0.019 & 0.985 &  & -0.002 & 0.020 & 0.020 & 0.960 \\
\hline
$\tau_{\NIE}(1)$ & $\mathsf{DeepMed}$    & -0.009 & 0.032 & 0.033 & 0.915 &  & 0.000  & 0.028 & 0.028 & 0.955 &  & 0.011  & 0.035 & 0.037 & 0.885 \\
                 & Lasso  & 0.059  & 0.078 & 0.098 & 0.890 &  & 0.058  & 0.093 & 0.110 & 0.920 &  & 0.118  & 0.042 & 0.125 & 0.270 \\
                 & RF     & 0.028  & 0.040 & 0.049 & 0.965 &  & 0.025  & 0.079 & 0.083 & 0.965 &  & 0.019  & 0.033 & 0.038 & 0.895 \\
                 & GBM    & 0.034  & 0.035 & 0.049 & 0.870 &  & 0.093  & 0.075 & 0.119 & 0.755 &  & 0.006  & 0.041 & 0.041 & 0.925 \\
                 & Oracle & 0.000  & 0.021 & 0.021 & 0.950 &  & 0.000  & 0.021 & 0.021 & 0.930 &  & 0.000  & 0.023 & 0.023 & 0.935 \\
\hline
$\tau_{\NIE}(0)$ & $\mathsf{DeepMed}$    & 0.002  & 0.037 & 0.037 & 0.915 &  & 0.002  & 0.027 & 0.027 & 0.940 &  & 0.008  & 0.031 & 0.032 & 0.935 \\
                 & Lasso  & 0.062  & 0.079 & 0.100 & 0.870 &  & 0.060  & 0.094 & 0.112 & 0.915 &  & 0.119  & 0.042 & 0.126 & 0.220 \\
                 & RF     & 0.002  & 0.039 & 0.039 & 0.980 &  & 0.025  & 0.078 & 0.082 & 0.965 &  & 0.012  & 0.032 & 0.034 & 0.920 \\
                 & GBM    & 0.033  & 0.035 & 0.048 & 0.855 &  & 0.088  & 0.075 & 0.116 & 0.795 &  & 0.012  & 0.040 & 0.042 & 0.935 \\
                 & Oracle & 0.000  & 0.021 & 0.021 & 0.940 &  & 0.000  & 0.022 & 0.022 & 0.920 &  & -0.001 & 0.023 & 0.023 & 0.925\\
\bottomrule[ 1pt]
\end{tabular}
}
\label{Table:p=20}
\end{table*}


\begin{table*}[hbt!]
\caption{The biases, empirical standard errors (SE) and root mean squared errors (RMSE) of the estimated total effects, NDE, and NIE, and the coverage probabilities (CP) of their corresponding $95 \%$ confidence intervals. There exist irrelevant covariates in this setup ($p = 100$). The simulation is based on 200 replicates.}  
\centering
 \resizebox{\textwidth}{60mm}{
\begin{tabular}{cccccccccccccccc}
\toprule[ 1pt]
              &        & \multicolumn{4}{c}{Case 1}        &  & \multicolumn{4}{c}{Case 2}       &  & \multicolumn{4}{c}{Case 3}       \\
\cline{3-6} \cline{8-11} \cline{13-16}
           & Method & Bias   & SE    & RMSE  & CP    &  & Bias   & SE    & RMSE  & CP    &  & Bias   & SE    & RMSE  & CP    \\
\midrule[ 1pt]
$\tau_{tot}$     & $\mathsf{DeepMed}$    & 0.008  & 0.046 & 0.047 & 0.955 &  & -0.018 & 0.033 & 0.038 & 0.895 &  & 0.019  & 0.043 & 0.047 & 0.915 \\
                 & Lasso  & 0.195  & 0.093 & 0.216 & 0.440 &  & -0.306 & 0.107 & 0.324 & 0.185 &  & 0.337  & 0.079 & 0.346 & 0.015 \\
                 & RF     & 0.187  & 0.052 & 0.194 & 0.095 &  & -0.205 & 0.095 & 0.226 & 0.560 &  & 0.206  & 0.057 & 0.214 & 0.040 \\
                 & GBM    & -0.016 & 0.035 & 0.038 & 0.965 &  & -0.155 & 0.072 & 0.171 & 0.440 &  & 0.082  & 0.053 & 0.098 & 0.620 \\
                 & Oracle & 0.000  & 0.030 & 0.030 & 0.945 &  & 0.000  & 0.030 & 0.030 & 0.960 &  & 0.001  & 0.031 & 0.031 & 0.935 \\
\hline
$\tau_{\NDE}(1)$ & $\mathsf{DeepMed}$    & 0.009  & 0.043 & 0.044 & 0.920 &  & -0.025 & 0.036 & 0.044 & 0.855 &  & 0.013  & 0.035 & 0.037 & 0.920 \\
                 & Lasso  & 0.129  & 0.050 & 0.138 & 0.275 &  & -0.369 & 0.055 & 0.373 & 0.000 &  & 0.215  & 0.067 & 0.225 & 0.095 \\
                 & RF     & 0.104  & 0.040 & 0.111 & 0.305 &  & -0.216 & 0.048 & 0.221 & 0.000 &  & 0.155  & 0.054 & 0.164 & 0.150 \\
                 & GBM    & -0.022 & 0.038 & 0.044 & 0.925 &  & -0.235 & 0.055 & 0.241 & 0.010 &  & 0.051  & 0.056 & 0.076 & 0.790 \\
                 & Oracle & 0.001  & 0.022 & 0.022 & 0.950 &  & 0.000  & 0.021 & 0.021 & 0.935 &  & 0.002  & 0.022 & 0.022 & 0.930 \\
\hline
$\tau_{\NDE}(0)$ & $\mathsf{DeepMed}$    & 0.006  & 0.040 & 0.040 & 0.920 &  & -0.018 & 0.036 & 0.040 & 0.855 &  & 0.016  & 0.033 & 0.037 & 0.925 \\
                 & Lasso  & 0.132  & 0.050 & 0.141 & 0.270 &  & -0.368 & 0.056 & 0.372 & 0.000 &  & 0.215  & 0.067 & 0.225 & 0.085 \\
                 & RF     & 0.063  & 0.038 & 0.074 & 0.690 &  & -0.213 & 0.047 & 0.218 & 0.005 &  & 0.139  & 0.056 & 0.150 & 0.255 \\
                 & GBM    & -0.032 & 0.037 & 0.049 & 0.870 &  & -0.241 & 0.054 & 0.247 & 0.005 &  & 0.059  & 0.057 & 0.082 & 0.765 \\
                 & Oracle & 0.001  & 0.022 & 0.022 & 0.940 &  & 0.000  & 0.021 & 0.021 & 0.935 &  & 0.002  & 0.022 & 0.022 & 0.940 \\
\hline
$\tau_{\NIE}(1)$ & $\mathsf{DeepMed}$    & 0.002  & 0.050 & 0.050 & 0.930 &  & 0.000  & 0.036 & 0.036 & 0.950 &  & 0.003  & 0.038 & 0.038 & 0.940 \\
                 & Lasso  & 0.063  & 0.080 & 0.102 & 0.885 &  & 0.062  & 0.089 & 0.108 & 0.915 &  & 0.122  & 0.044 & 0.130 & 0.200 \\
                 & RF     & 0.124  & 0.050 & 0.134 & 0.335 &  & 0.008  & 0.099 & 0.099 & 0.945 &  & 0.068  & 0.036 & 0.077 & 0.395 \\
                 & GBM    & 0.016  & 0.040 & 0.043 & 0.905 &  & 0.086  & 0.084 & 0.120 & 0.835 &  & 0.023  & 0.036 & 0.043 & 0.905 \\
                 & Oracle & -0.001 & 0.021 & 0.021 & 0.930 &  & 0.000  & 0.021 & 0.021 & 0.950 &  & -0.001 & 0.021 & 0.021 & 0.950 \\
\hline
$\tau_{\NIE}(0)$ & $\mathsf{DeepMed}$    & -0.001 & 0.044 & 0.044 & 0.945 &  & 0.007  & 0.037 & 0.038 & 0.920 &  & 0.006  & 0.036 & 0.036 & 0.960 \\
                 & Lasso  & 0.066  & 0.081 & 0.104 & 0.870 &  & 0.063  & 0.090 & 0.110 & 0.915 &  & 0.123  & 0.044 & 0.131 & 0.215 \\
                 & RF     & 0.084  & 0.047 & 0.096 & 0.645 &  & 0.011  & 0.098 & 0.099 & 0.945 &  & 0.051  & 0.034 & 0.061 & 0.615 \\
                 & GBM    & 0.007  & 0.038 & 0.039 & 0.940 &  & 0.080  & 0.082 & 0.115 & 0.845 &  & 0.031  & 0.034 & 0.046 & 0.890 \\
                 & Oracle & -0.001 & 0.021 & 0.021 & 0.935 &  & 0.000  & 0.021 & 0.021 & 0.935 &  & -0.001 & 0.021 & 0.021 & 0.960\\
\bottomrule[ 1pt]
\end{tabular}
}

\label{Table:p=100}
\end{table*}


\begin{table*}[hbt!]
\caption{The validation loss of the nuisance functions. The cross-entropy loss is used for fitting $a (d | x, m)$ and $a (d | x)$, and mean squared loss is used for fitting the other nuisance functions. There exist no irrelevant covariates in this setup.} 
\centering 
\begin{tabular}{m{1cm}<{\centering}m{1cm}<{\centering}m{1.3cm}<{\centering}m{1.5cm}<{\centering}m{1.3cm}<{\centering}m{1.3cm}<{\centering}m{1.3cm}<{\centering}m{1.3cm}<{\centering}m{1.5cm}<{\centering}m{1.3cm}<{\centering}}
\toprule[ 1pt]
& & \small{$a (1 | x, m)$} & \small{$a (1 | x)$} &\small{$\mu (x, 1, m)$} & \small{$\E_{0} (\mu_1)^*$} &  \small{$\mu (x, 1)$} & \small{$\mu (x, 0, m)$} &    \small{$\E_{1} (\mu_0)^*$} &  \small{$\mu (x, 0)$}  \\

\midrule[ 1pt]       
Case 1 & $\mathsf{DeepMed}$    & 0.646 & 0.647 & 1.151 & 1.353  & 2.290  & 1.172 & 1.275  & 2.304  \\
       & Lasso  & 0.660 & 0.660 & 5.677 & 14.889 & 20.725 & 5.634 & 15.099 & 20.705 \\
       & RF     & 0.662 & 0.664 & 3.284 & 6.322  & 6.189  & 3.777 & 5.618  & 6.909  \\
       & GBM    & 0.651 & 0.651 & 2.344 & 3.003  & 3.290  & 2.383 & 2.819  & 3.370  \\
       & Oracle & 0.639 & 0.642 & 1.057 & 1.031  & 2.100  & 1.063 & 1.033  & 2.108  \\
       \hline
Case 2 & $\mathsf{DeepMed}$    & 0.680 & 0.681 & 1.309 & 1.434  & 2.318  & 1.305 & 1.213  & 2.311  \\
       & Lasso  & 0.694 & 0.694 & 8.285 & 23.037 & 31.239 & 8.275 & 23.117 & 31.278 \\
       & RF     & 0.694 & 0.697 & 5.046 & 19.915 & 16.393 & 4.924 & 19.291 & 16.568 \\
       & GBM    & 0.688 & 0.689 & 4.587 & 11.961 & 8.089  & 4.441 & 11.592 & 7.918  \\
       & Oracle & 0.670 & 0.676 & 1.055 & 1.037  & 2.109  & 1.061 & 1.039  & 2.116  \\
        \hline
Case 3 & $\mathsf{DeepMed}$    & 0.647 & 0.649 & 1.35  & 1.388 & 2.572  & 1.376 & 1.36  & 2.568 \\
       & Lasso  & 0.662 & 0.664 & 9.31  & 4.612 & 13.935 & 8.945 & 4.584 & 13.35 \\
       & RF     & 0.657 & 0.664 & 4.966 & 2.728 & 5.388  & 5.101 & 2.647 & 5.425 \\
       & GBM    & 0.648 & 0.649 & 4.087 & 3.136 & 4.162  & 4.131 & 2.946 & 4.172 \\
       & Oracle & 0.637 & 0.643 & 1.031 & 1.019 & 2.11   & 1.033 & 1.02  & 2.111 \\
\bottomrule[ 1pt]
\end{tabular}
$^*\E_{0} (\mu_1) = \E [\mu (X, D = 1, M) | X, D = 0]$ and $\E_{1} (\mu_0) = \E [\mu (X, D = 0, M) | X, D = 1]$.
\label{Table: cv loss/p=5}
\end{table*}


\begin{table*}[hbt!]
\caption{The validation loss of the nuisance functions. The cross-entropy loss is used for fitting $a (d | x, m)$ and $a (d | x)$, and mean squared loss is used for fitting the other nuisance functions. There exist irrelevant covariates in this setup ($p = 20$).}  
\centering
\begin{tabular}{m{1cm}<{\centering}m{1cm}<{\centering}m{1.3cm}<{\centering}m{1.5cm}<{\centering}m{1.3cm}<{\centering}m{1.3cm}<{\centering}m{1.3cm}<{\centering}m{1.3cm}<{\centering}m{1.5cm}<{\centering}m{1.3cm}<{\centering}}
\toprule[ 1pt]
& & \small{$a (1 | x, m)$} & \small{$a (1 | x)$} &\small{$\mu (x, 1, m)$} & \small{$\E_{0} (\mu_1)^*$} & \small{$\mu (1, x)$} & \small{$\mu (x, 0, m)$} & \small{$\E_{1} (\mu_0)^*$} & \small{$\mu (0, x)$}  \\

\midrule[ 1pt]       
Case 1 & $\mathsf{DeepMed}$    & 0.657 & 0.659 & 1.316 & 2.943  & 3.241  & 1.454 & 2.709  & 3.319  \\
       & Lasso  & 0.666 & 0.667 & 5.583 & 14.84  & 20.663 & 5.541 & 15.25  & 20.65  \\
       & RF     & 0.662 & 0.661 & 4.449 & 8.407  & 8.078  & 4.958 & 7.619  & 8.628  \\
       & GBM    & 0.657 & 0.658 & 2.987 & 3.24   & 3.976  & 3.11  & 3.265  & 3.992  \\
       & Oracle & 0.64  & 0.645 & 1.014 & 1.019  & 2.097  & 1.009 & 1.02   & 2.087  \\
       \hline
Case 2 & $\mathsf{DeepMed}$    & 0.692 & 0.691 & 1.42  & 1.822  & 2.432  & 1.543 & 1.793  & 2.533  \\
       & Lasso  & 0.694 & 0.694 & 7.98  & 22.882 & 31.023 & 8.003 & 23.34  & 30.977 \\
       & RF     & 0.692 & 0.692 & 5.708 & 25.493 & 22.644 & 5.618 & 26.092 & 21.803 \\
       & GBM    & 0.693 & 0.693 & 5.176 & 17.787 & 12.627 & 5.187 & 18.722 & 12.227 \\
       & Oracle & 0.672 & 0.677 & 1.014 & 1.017  & 2.095  & 1.01  & 1.018  & 2.083  \\
       \hline
Case 3 & $\mathsf{DeepMed}$    & 0.653 & 0.652 & 1.448 & 2.456 & 3.214  & 1.504 & 2.512 & 3.172  \\
       & Lasso  & 0.655 & 0.657 & 9.581 & 4.161 & 13.785 & 8.699 & 4.663 & 13.162 \\
       & RF     & 0.649 & 0.651 & 6.626 & 2.824 & 6.42   & 6.789 & 2.924 & 6.591  \\
       & GBM    & 0.646 & 0.647 & 6.344 & 3.372 & 5.14   & 6.49  & 3.619 & 5.426  \\
       & Oracle & 0.635 & 0.638 & 1.021 & 1.034 & 2.059  & 1.025 & 1.037 & 2.061  \\
\bottomrule[ 1pt]
\end{tabular}
$^*\E_{0} (\mu_1) = \E [\mu (X, D = 1, M) | X, D = 0]$ and $\E_{1} (\mu_0) = \E [\mu (X, D = 0, M) | X, D = 1]$.
\label{Table: cv loss/p=20}
\end{table*}


\begin{table*}[hbt!]
\caption{The validation loss of the nuisance functions. The cross-entropy loss is used for fitting $a (d | x, m)$ and $a (d | x)$, and mean squared loss is used for fitting the other nuisance functions. There exist irrelevant covariates in this setup ($p = 100$).}  
\centering
\begin{tabular}{m{1cm}<{\centering}m{1cm}<{\centering}m{1.3cm}<{\centering}m{1.5cm}<{\centering}m{1.3cm}<{\centering}m{1.3cm}<{\centering}m{1.3cm}<{\centering}m{1.3cm}<{\centering}m{1.5cm}<{\centering}m{1.3cm}<{\centering}}
\toprule[ 1pt]
& & \small{$a (1 | x, m)$} & \small{$a (1 | x)$} & \small{$\mu (x, 1, m)$} & \small{$\E_{0} (\mu_1)^*$} & \small{$\mu (x, 1)$} & \small{$\mu (x, 0, m)$} & \small{$\E_{1} (\mu_0)^*$} & \small{$\mu (x, 0)$} \\

\midrule[ 1pt]       
Case 1 & $\mathsf{DeepMed}$    & 0.672 & 0.672 & 1.769 & 4.154  & 4.071  & 1.692 & 4.723  & 4.138  \\
       & Lasso  & 0.667 & 0.667 & 5.438 & 14.828 & 20.607 & 5.478 & 15.592 & 20.607 \\
       & RF     & 0.667 & 0.668 & 4.964 & 9.123  & 8.836  & 5.487 & 8.794  & 9.530   \\
       & GBM    & 0.664 & 0.663 & 3.266 & 3.659  & 4.112  & 3.469 & 3.684  & 4.147  \\
       & Oracle & 0.648 & 0.652 & 1.001 & 1.025  & 2.069  & 1.001 & 1.027  & 2.066  \\
       \hline
Case 2 & $\mathsf{DeepMed}$    & 0.695 & 0.694 & 1.81  & 2.583  & 2.909  & 1.926 & 2.615  & 2.936  \\
       & Lasso  & 0.695 & 0.695 & 8.006 & 22.765 & 31.272 & 8.005 & 23.63  & 31.239 \\
       & RF     & 0.692 & 0.692 & 5.708 & 25.493 & 22.644 & 5.618 & 26.092 & 21.803 \\
       & GBM    & 0.688 & 0.689 & 4.587 & 11.961 & 8.089  & 4.441 & 11.592 & 7.918  \\
       & Oracle & 0.677 & 0.68  & 1.003 & 1.022  & 2.067  & 1.003 & 1.023  & 2.065  \\
       \hline
Case 3 & $\mathsf{DeepMed}$    & 0.679 & 0.677 & 2.416 & 4.563 & 4.643  & 2.500   & 5.098 & 4.621 \\
       & Lasso  & 0.660  & 0.662 & 9.784 & 4.858 & 14.364 & 8.523 & 4.360  & 13.000    \\
       & RF     & 0.661 & 0.663 & 8.124 & 3.102 & 7.808  & 7.973 & 2.970  & 7.586 \\
       & GBM    & 0.653 & 0.653 & 7.906 & 3.320  & 6.205  & 7.252 & 3.101 & 6.159 \\
       & Oracle & 0.635 & 0.641 & 1.011 & 1.039 & 2.144  & 1.014 & 1.042 & 2.144  \\
\bottomrule[ 1pt]
\end{tabular}
$^*\E_{0} (\mu_1) = \E [\mu (X, D = 1, M) | X, D = 0]$ and $\E_{1} (\mu_0) = \E [\mu (X, D = 0, M) | X, D = 1]$.
\label{Table: cv loss/p=100}
\end{table*}

\begin{table*}[hbt!]
\centering
\caption{The validation losses of nuisance functions in real data application to the COMPAS algorithm fairness.}

\begin{tabular}{lcccc}
\toprule[ 1pt]
                          & $\mathsf{DeepMed}$   & Lasso & RF    & GBM   \\
\midrule[ 1pt]
$a (1 | x, m)$         & 0.622 & 0.626 & 0.638 & 0.625 \\
$a (1 | x)$           & 0.648 & 0.650 & 0.699 & 0.650 \\
$\mu (x, 1, m)$         & 4.924 & 5.480 & 5.436 & 5.064 \\
$\E [\mu (X, D = 1, M) | X = x, D = 0]$ & 1.636 & 0.993 & 0.832 & 1.579 \\
$\mu (x, 1)$           & 7.265 & 7.392 & 7.393 & 7.378 \\
$\mu (x, 0, m)$         & 3.710 & 4.108 & 4.266 & 3.928 \\
$\E [\mu (X, D = 0, M) | X = x, D = 1]$ & 7.582 & 2.443 & 0.993 & 2.012 \\
$\mu (x, 0)$           & 5.197 & 5.299 & 5.414 & 5.269 \\

\bottomrule[ 1pt]
\end{tabular}
\label{Table real2:validation loss}
\end{table*}

\newpage
\section{Additional information on synthetic experiments}\label{app:sim}
In all the synthetic experiments, we adopt a 3-fold cross-validation to choose the hyperparameters of DNN, RF and GBM over a grid of candidate values. For DNN, we fix the batch-size as 100, and choose depth $L$ from 1 to 3, width $K$ from 10 to 500, $L_1$-regularization parameter $\lambda$ from 0 to 0.4, and epochs from 100 to 500.  For RF, we choose the number of trees from 1 to 20, and maximum number of nodes from 10 to 1000. For GBM, we choose the number of trees from 1 to 20, and depth from 10 to 1000. The other hyperparameters are set to the default values in the $\mathsf{R}$ packages. As mentioned in the main text, we leave its theoretical justification and other alternative approaches such as the minimax criterion \citepSM{robins2020double, cui2019selective} or CTMLE (\citetSM{van2010collaborative}; also see Chapter 2 of \citetSM{liu2018contributions}) to future works.

Case 4 (H\"{o}lder functions): we repeat the simulation in Case 3 but set $\alpha = 0.6$ to further decrease the smoothness of the H\"{o}lder functions. In particular, $\alpha = 0.6$ is close to the limit ($\alpha = 0.5 + \epsilon$ for arbitrarily small $\epsilon > 0$) for $\mathsf{DeepMed}$ estimators to be semiparametric efficient based on Theorem \ref{thm:holder}. Thus we can examine whether surpassing this limit for nuisance function estimates computed by ERM \eqref{erm} without considering the DNN training process actually translates to practical success of $\mathsf{DeepMed}$. Unfortunately, the results in Table \ref{Table case4} show otherwise. In general, $\mathsf{DeepMed}$ still has superior performance than the other competing methods for $p = 5, 20, 100$. However, even at $p = 5$, the biases of the $\mathsf{DeepMed}$ estimators are close to their standard errors. As a result, their CPs undercover the true causal parameters (though the CP is not very far from 95\%). However, based on Lemma \ref{lem:holder}, one should be able to estimate all the nuisance functions at rate $O (n^{- 1 / 4})$ as $\alpha > 0.5$ if the nuisance function estimates are solutions to ERM \eqref{erm}, which should in turn leads to the semiparametric efficiency of $\mathsf{DeepMed}$ NDE/NIE estimators and valid inference.

There are several possible explanations for the $\mathsf{DeepMed}$ estimators failing to be semiparametric efficient: (1) it is entirely possible that gradient-based training algorithms such as $\mathsf{adam}$ (used in our paper) or SGD could find nuisance function estimators from the FNN-ReLU class that are close to the ERM \eqref{erm} but we just failed to do so in our implementation; or (2) it is a manifestation of a certain ``low-frequency-bias'' \citepSM{rahaman2019spectral, hu2020surprising, xu2020frequency} of DNNs trained by gradient-based algorithms or the computational hardness of learning DNNs by gradient-based algorithms \citepSM{goel2020superpolynomial, chen2022learning}, which predicts that DNNs trained by off-the-shelf algorithms bias towards functions with lower complexity. We conjecture that (2) is indeed the reason. It suggests that in future works, to fully establish the practically relevant statistical guarantees of DNN-based nonparametric regression or DNN-based causal inference method such as $\mathsf{DeepMed}$, it is important to take the effect of training algorithms into consideration.

\begin{remark}\label{rem:barron}
As we have discussed, most established convergence rates of DNNs in the nonparametric statistics literature do not take the potential implicit regularization effect or bias of the training algorithms into account. In this remark, we suggest several possible directions to explore in future works. The so-called Barron space \citepSM{e2021barron} was recently shown to be a natural function space describing the class of neural network functions trained with SGD. If the nuisance functions lie in a Barron space, then the rate of convergence is dimension-independent \citepSM{e2021barron, chen2021representation}, which seems to be consistent with the observation that DNNs overcome curse-of-dimensionality in practice. However, as shown in \citetSM{e2021barron}, the complexity of Barron functions is extremely small, thus casting doubt on if the theoretical claim under Barron spaces is ``too good to be true'' in fields such as biomedical and social sciences or algorithmic fairness, in which model misspecification bias might have catastrophic consequences. Recent results \citepSM{siegel2021sharp} trying to generalize Barron spaces to model more complex functions might be a useful direction to consider in problems related to semiparametric causal inference.
\end{remark}

Case 5 (composition H\"{o}lder functions): In the last setting, we consider composition H\"{o}lder functions by composing $\eta (x; \alpha) \in \mathcal{H}_{1} (\alpha; B)$ hierarchically for some constant $B > 0$ as follows:
\begin{align*}
    d (x) & = 0.2 \eta \left( \sum_{i = 1}^3 x_i; \alpha \right) + 0.2 \eta \left( \sum_{i = 1}^3 \eta (x_i; \alpha); \alpha \right), \\
    m (x) & = 0.5 \eta \left( \sum_{i = 1}^3 x_i; \alpha \right) + 0.2 \eta \left( \sum_{i = 1}^3 \eta (x_i; \alpha); \alpha \right), \\
    y (x) & = 0.2 \eta \left( \sum_{i = 1}^3 x_i; \alpha \right) + 0.5 \eta \left( \sum_{i = 1}^3 \eta (x_i; \alpha); \alpha \right),
\end{align*} 
where we set $\alpha = 1.5$. We choose the ``depth'' of compositions as 2 for simplicity. The nuisance functions $a, f, \mu$ in Case 5 correspond to the composition H\"{o}lder functions studied in the seminal work by \citetSM{schmidt2020nonparametric}. For such function spaces, \citetSM{schmidt2020nonparametric} showed that linear estimators cannot achieve minimax optimal estimation rate, yet nonlinear DNN-based estimators can. As shown in Table \ref{Table case5}, the $\mathsf{DeepMed}$ estimators do exhibit superior performance compared with the other competing methods but they are far from being semiparametric efficient. When $\alpha = 1.5$, at least based on \citetSM{schmidt2020nonparametric}, the ERM-based DNN regression estimators should converge to the true function in $L_{2}$-norm at a rate faster than $n^{- 1 / 4}$. Then using Proposition \ref{thm:master}, the $\mathsf{DeepMed}$ estimators should have been semiparametric efficient. But as in Table \ref{Table case5}, our empirical results suggest otherwise. This is another instance that suggests the necessity of developing more refined theoretical properties of $\mathsf{DeepMed}$.

In results shown previously, $\mathsf{adam}$ \citepSM{kingma2015adam} was used to train the DNN weights. Finally, in Table \ref{Table:SGD}, we also display the simulation results when DNN weights were trained by vanilla SGD. Again, as expected, the $\mathsf{DeepMed}$ causal effect estimates with DNN weights trained by SGD are not semiparametric efficient.

\begin{remark}
$\mathsf{DeepMed}$ has the option of estimating nuisance functions by other types of machine learning methods, such as those mentioned in Section \ref{sec:sim}. It is important to develop statistical methodology that can help practitioners decide which method one should use, especially when different methods output qualitatively different results. This is an important research question to pursue as mediation analysis is often followed with critical decision making; a recent proposal can be found in \citetSM{liu2020nearly}.
\end{remark}

\begin{table*}[hbt!] 
\caption{Simulation Case 4 ($\alpha=0.6$): The biases, empirical standard errors (SE) and root mean squared errors (RMSE) of the estimated average treatment effects and coverage probabilities (CP) of 95$\%$ confidence intervals. The simulation is based on 200 replicates.}  
\centering
 \resizebox{\textwidth}{60mm}{
\begin{tabular}{cccccccccccccccc}
\toprule[ 1pt]
              &        & \multicolumn{4}{c}{$p=5$}        &  & \multicolumn{4}{c}{$p=20$}       &  & \multicolumn{4}{c}{$p=100$}       \\
\cline{3-6} \cline{8-11} \cline{13-16}
           & Method & Bias   & SE    & RMSE  & CP    &  & Bias   & SE    & RMSE  & CP    &  & Bias   & SE    & RMSE  & CP    \\
\midrule[ 1pt]
$\tau_{tot}$& $\mathsf{DeepMed}$ & 0.027 & 0.047 & 0.054 & 0.880 &  & 0.035 & 0.047 & 0.059 & 0.880 &  & 0.062 & 0.051 & 0.080 & 0.740  \\ 
&        Lasso & 0.353 & 0.087 & 0.364 & 0.010 &  & 0.335 & 0.082 & 0.345 & 0.020 &  & 0.343 & 0.087 & 0.354 & 0.020  \\ 
&        RF & 0.004 & 0.051 & 0.051 & 0.995 &  & 0.080 & 0.058 & 0.099 & 0.780 &  & 0.202 & 0.065 & 0.212 & 0.095  \\ 
&       GBM & 0.020 & 0.050 & 0.054 & 0.930 &  & 0.055 & 0.056 & 0.078 & 0.830 &  & 0.088 & 0.061 & 0.107 & 0.650  \\ 
&        Oracle & -0.002 & 0.031 & 0.031 & 0.955 &  & -0.003 & 0.031 & 0.031 & 0.950 &  & 0.002 & 0.031 & 0.031 & 0.930  \\ 
 \hline
$\tau_{\NDE}(1)$& $\mathsf{DeepMed}$ & 0.004 & 0.038 & 0.038 & 0.925 &  & -0.006 & 0.041 & 0.041 & 0.940 &  & 0.003 & 0.047 & 0.047 & 0.945  \\
&        Lasso & 0.227 & 0.071 & 0.238 & 0.095 &  & 0.211 & 0.066 & 0.221 & 0.120 &  & 0.217 & 0.073 & 0.229 & 0.115  \\ 
&        RF & 0.022 & 0.045 & 0.050 & 0.980 &  & 0.087 & 0.054 & 0.102 & 0.690 &  & 0.175 & 0.060 & 0.185 & 0.140  \\ 
&        GBM & 0.017 & 0.051 & 0.054 & 0.905 &  & 0.044 & 0.057 & 0.072 & 0.845 &  & 0.073 & 0.062 & 0.096 & 0.750  \\ 
&        Oracle & 0.000 & 0.022 & 0.022 & 0.965 &  & -0.003 & 0.022 & 0.022 & 0.920 &  & 0.002 & 0.022 & 0.022 & 0.940  \\ 
\hline
$\tau_{\NDE}(0)$&$\mathsf{DeepMed}$ & 0.005 & 0.034 & 0.034 & 0.945 &  & 0.005 & 0.038 & 0.038 & 0.960 &  & 0.014 & 0.047 & 0.049 & 0.925  \\
&        Lasso & 0.229 & 0.069 & 0.239 & 0.090 &  & 0.213 & 0.066 & 0.223 & 0.115 &  & 0.217 & 0.073 & 0.229 & 0.125  \\ 
&        RF & 0.032 & 0.047 & 0.057 & 0.960 &  & 0.081 & 0.051 & 0.096 & 0.730 &  & 0.158 & 0.061 & 0.169 & 0.210  \\ 
&        GBM & 0.023 & 0.050 & 0.055 & 0.930 &  & 0.049 & 0.056 & 0.074 & 0.865 &  & 0.075 & 0.061 & 0.097 & 0.715  \\ 
&        Oracle & 0.000 & 0.022 & 0.022 & 0.965 &  & -0.003 & 0.023 & 0.023 & 0.925 &  & 0.002 & 0.021 & 0.021 & 0.950  \\ 
 \hline
$\tau_{\NIE}(1)$&$\mathsf{DeepMed}$ & 0.021 & 0.039 & 0.044 & 0.865 &  & 0.030 & 0.042 & 0.052 & 0.865 &  & 0.047 & 0.044 & 0.064 & 0.830  \\ 
&        Lasso & 0.124 & 0.051 & 0.134 & 0.295 &  & 0.123 & 0.048 & 0.132 & 0.280 &  & 0.126 & 0.049 & 0.135 & 0.260  \\ 
&        RF & -0.029 & 0.040 & 0.049 & 0.895 &  & -0.002 & 0.039 & 0.039 & 0.935 &  & 0.044 & 0.038 & 0.058 & 0.760  \\ 
&        GBM & -0.003 & 0.042 & 0.042 & 0.935 &  & 0.006 & 0.045 & 0.045 & 0.905 &  & 0.013 & 0.039 & 0.041 & 0.935  \\ 
&        Oracle & -0.002 & 0.024 & 0.024 & 0.940 &  & 0.000 & 0.023 & 0.023 & 0.930 &  & -0.001 & 0.023 & 0.023 & 0.950  \\ 
        \hline
$\tau_{\NIE}(0)$&$\mathsf{DeepMed}$ & 0.023 & 0.040 & 0.046 & 0.875 &  & 0.041 & 0.041 & 0.058 & 0.815 &  & 0.059 & 0.047 & 0.075 & 0.730  \\ 
&        Lasso & 0.126 & 0.052 & 0.136 & 0.270 &  & 0.124 & 0.048 & 0.133 & 0.250 &  & 0.126 & 0.049 & 0.135 & 0.250  \\ 
&        RF & -0.019 & 0.042 & 0.046 & 0.935 &  & -0.007 & 0.035 & 0.036 & 0.955 &  & 0.027 & 0.036 & 0.045 & 0.885  \\ 
&        GBM & 0.004 & 0.042 & 0.042 & 0.920 &  & 0.012 & 0.041 & 0.043 & 0.930 &  & 0.015 & 0.039 & 0.042 & 0.930  \\ 
&        Oracle & -0.001 & 0.024 & 0.024 & 0.945 &  & 0.000 & 0.023 & 0.023 & 0.925 &  & -0.001 & 0.023 & 0.023 & 0.955\\
\bottomrule[ 1pt]
\end{tabular}
}
\label{Table case4}
\end{table*}

\begin{table*}[hbt!]
\centering
\caption{Simulation Case 5: The biases, empirical standard errors (SE) and root mean squared errors (RMSE) of the estimated average treatment effects and coverage probabilities (CP) of 95$\%$ confidence intervals. The simulation is based on 200 replicates.}  
\begin{tabular}{llllll}
\toprule
                 & Method & Bias   & SE    & RMSE  & CP    \\
\midrule                
$\tau_{tot}$     & $\mathsf{DeepMed}$    & 0.225  & 0.032 & 0.227 & 0.000 \\
                 & Lasso  & 0.369  & 0.032 & 0.370 & 0.000     \\
                 & RF     & 0.277  & 0.035 & 0.279 & 0.000 \\
                 & GBM    & 0.368  & 0.032 & 0.369 & 0.000 \\
                 & Oracle & -0.003 & 0.028 & 0.028 & 0.960 \\
\hline
$\tau_{\NDE} (1)$ & $\mathsf{DeepMed}$    & 0.040  & 0.023 & 0.046 & 0.570 \\
                 & Lasso  & 0.113  & 0.023 & 0.115 & 0.005 \\
                 & RF     & 0.087  & 0.029 & 0.092 & 0.170 \\
                 & GBM    & 0.114  & 0.024 & 0.116 & 0.005 \\
                 & Oracle & -0.001 & 0.020 & 0.020 & 0.965 \\
\hline
$\tau_{\NDE} (0)$ & $\mathsf{DeepMed}$    & 0.044  & 0.023 & 0.050 & 0.470 \\
                 & Lasso  & 0.112  & 0.023 & 0.114 & 0.005 \\
                 & RF     & 0.083  & 0.029 & 0.088 & 0.200 \\
                 & GBM    & 0.114  & 0.024 & 0.116 & 0.005 \\
                 & Oracle & -0.002 & 0.020 & 0.020 & 0.960 \\
\hline
$\tau_{\NIE} (1)$ & $\mathsf{DeepMed}$    & 0.181  & 0.027 & 0.183 & 0.000 \\
                 & Lasso  & 0.257  & 0.026 & 0.258 & 0.000 \\
                 & RF     & 0.193  & 0.033 & 0.196 & 0.000 \\
                 & GBM    & 0.255  & 0.027 & 0.256 & 0.000 \\
                 & Oracle & -0.001 & 0.021 & 0.021 & 0.955 \\
\hline
$\tau_{\NIE} (0)$ & $\mathsf{DeepMed}$    & 0.184  & 0.028 & 0.186 & 0.000 \\
                 & Lasso  & 0.256  & 0.027 & 0.257 & 0.000 \\
                 & RF     & 0.190  & 0.034 & 0.193 & 0.000 \\
                 & GBM    & 0.255  & 0.027 & 0.256 & 0.000 \\
                 & Oracle & -0.002 & 0.021 & 0.021 & 0.940 \\
\bottomrule
\end{tabular}
\label{Table case5}
\end{table*}


\begin{table*}[hbt!]
\centering
\caption{The simulation results under Cases 4-5 for DeepMed with DNN weights trained by SGD. The biases, empirical standard errors (SE) and root mean squared errors (RMSE) of the estimated average treatment effects and coverage probabilities (CP) of 95$\%$ confidence intervals. The simulation is based on 200 replicates.}  
\begin{tabular}{lllllll}
\toprule
              &   &  & Bias   & SE    & RMSE  & CP    \\
\midrule                
Case 4 & $p=5$   & $\tau_{tot}$      & 0.033 & 0.049 & 0.059 & 0.870 \\
       &       & $\tau_{\NDE} (1)$ & 0.007 & 0.037 & 0.038 & 0.955 \\
       &       & $\tau_{\NDE} (0)$ & 0.004 & 0.039 & 0.039 & 0.940 \\
       &       & $\tau_{\NIE} (1)$ & 0.028 & 0.041 & 0.050 & 0.825 \\
       &       & $\tau_{\NIE} (0)$ & 0.026 & 0.040 & 0.048 & 0.815 \\
\hline
       & $p=20$  & $\tau_{tot}$      & 0.060 & 0.050 & 0.078 & 0.770 \\
       &       & $\tau_{\NDE} (1)$ & 0.006 & 0.045 & 0.045 & 0.925 \\
       &       & $\tau_{\NDE} (0)$ & 0.011 & 0.044 & 0.045 & 0.910 \\
       &       & $\tau_{\NIE} (1)$ & 0.049 & 0.043 & 0.065 & 0.710 \\
       &       & $\tau_{\NIE} (0)$ & 0.054 & 0.048 & 0.072 & 0.715 \\
\hline 
       & $p=100$ & $\tau_{tot}$      & 0.073 & 0.050 & 0.088 & 0.655 \\
       &       & $\tau_{\NDE} (1)$ & 0.026 & 0.059 & 0.064 & 0.835 \\
       &       & $\tau_{\NDE} (0)$ & 0.027 & 0.055 & 0.061 & 0.865 \\
       &       & $\tau_{\NIE} (1)$ & 0.046 & 0.051 & 0.069 & 0.775 \\
       &       & $\tau_{\NIE} (0)$ & 0.047 & 0.059 & 0.075 & 0.745 \\
\hline
Case 5 &     & $\tau_{tot}$      & 0.229 & 0.034 & 0.232 & 0.000 \\
       &       & $\tau_{\NDE} (1)$ & 0.043 & 0.024 & 0.049 & 0.495 \\
       &       & $\tau_{\NDE} (0)$ & 0.040 & 0.024 & 0.047 & 0.590 \\
       &       & $\tau_{\NIE} (1)$ & 0.190 & 0.027 & 0.192 & 0.000 \\
       &       & $\tau_{\NIE} (0)$ & 0.186 & 0.027 & 0.188 & 0.000 \\
\bottomrule
\end{tabular}
\label{Table:SGD}
\end{table*}

\section{A comment on the potential issue of unmeasured confounding in applications related to algorithmic fairness}
\label{app:endogeneity}
Finally, we briefly comment on the potential issue of unmeasured confounding in applications related to algorithmic fairness. It is definitely possible to have unmeasured treatment-mediator confounding in real data analysis. But unmeasured treatment-outcome and mediator-outcome confounding may not be huge issues in mediation analysis related to algorithmic fairness because in most cases we have access to all the features used to fit the prediction map, which is the outcome $Y$ in our notation. However, they could be violated when some of the features used to fit the model are concealed to protect data privacy. We did not consider these issues in the real data application but we admit that the final results might be biased due to these unmeasured confounding biases. But the same caveat also applies to most of the other works using mediation analysis in algorithmic fairness. 

For fields such as epidemiology or social sciences, more often than not, we do not have the luxury of having access to all important features for the mediator, exposure, and outcome. Thus in general, it is important to incorporate instrumental variables \citepSM{frolich2017direct}, valid proxies \citepSM{dukes2021proximal} or other identification strategies \citepSM{sun2022semiparametric} into $\mathsf{DeepMed}$ to handle unmeasured confounding.

\section{Real data analysis (continued)}
\label{app:real}
This section is a continuation of Section \ref{sec:real} of the main text.

In this section, we apply $\mathsf{DeepMed}$ to a second dataset and study whether gender has direct effect on personal annual income not mediated by occupation. We use the Adult dataset (\href{https://archive.ics.uci.edu/ml/datasets/adult}{https://archive.ics.uci.edu/ml/datasets/adult}) from the 1994 Census database in U.S., which includes 48,842 individuals \citepSM{kohavi1996scaling}. We set $D = 1$ for male and $D = 0$ for female. Occupation ($M$) is a categorical variable containing 14 general types of occupations. The personal annual income is a binary variable, with $Y = 1$ (or $Y = 0$) indicating that an individual makes more (or less) than \$50,000 annually. We also include age, race, education level and employment status as covariates. After removing observations with missing values, the remaining sample size is 45,997.

In this example, since $M$ is multi-dimensional, we utilize the alternative parameterization strategy described in Remark \ref{rem:bayes} and estimate the propensity scores $a (d | x)$ and $a (d | x, m)$ before and after conditioning on the mediators $M$, together with regressing $\hat{\mu} (x, d, m)$ against $(x, d)$, all using DNNs. One may be concerned with the potential incoherence between the posited models for the propensity scores $a (d | x, m)$ and $a (d | x)$ and the joint distribution of the observed data $(X, A, M, Y)$. This incoherence could be problematic when parametric models are posited. However, under the semiparametric framework, it is of secondary concern to correctly model the joint distribution of the observed data, which is indeed a very difficult problem. More emphasis is put on how well the target causal parameters such as NDE/NIE are estimated. As long as the nonparametric estimates $\hat{a} (d | x)$ and $\hat{a} (d | x, m)$ converge to the true nuisance functions at sufficiently fast rates, the estimates of the target causal parameters should be sufficiently accurate. 

All the methods find significant NDE of gender on personal annual income (see Table \ref{Table real3}). This positive NDE suggests that males tend to have higher income than females, and this cannot be explained by the indirect effect through occupation. In this dataset, the $\mathsf{DeepMed}$ estimators again have smaller validation errors than the other competing methods, possibly suggesting smaller biases of the corresponding NDE/NIE estimators (see Table \ref{Table real3:validation loss}).

\begin{table}[hbt!]
\caption{Real data application to income fairness. The estimated NDE/NIE of gender ($D$) on income ($Y$) with occupation ($M$) as the mediator.}  
\centering
\begin{tabular}{cccccccccccccccc}
\toprule[ 1pt]
  
          Method & Effect & Estimate   & SE    & P value     \\
\midrule[ 1pt]
$\mathsf{DeepMed}$   & $\tau_{tot}$    & 0.155 & 0.004 & $< 10^{-16}$ \\ 
      & $\tau_{\NDE}(1)$ & 0.161 & 0.007 & $< 10^{-16}$ \\ 
      & $\tau_{\NDE}(0)$ & 0.148 & 0.004 & $< 10^{-16}$ \\ 
      & $\tau_{\NIE}(1)$ & 0.007 & 0.002 & 0.003 \\
      & $\tau_{\NIE}(0)$ & -0.005 & 0.005 & 0.343 \\
      \hline
Lasso & $\tau_{tot}$    & 0.171 & 0.004 & $< 10^{-16}$ \\ 
      & $\tau_{\NDE}(1)$ & 0.165 & 0.006 & $< 10^{-16}$ \\ 
      & $\tau_{\NDE}(0)$ & 0.155 & 0.004 & $< 10^{-16}$ \\ 
      & $\tau_{\NIE}(1)$ & 0.016 & 0.002 & $3 \times 10^{-11}$ \\ 
      & $\tau_{\NIE}(0)$ & 0.006 & 0.004 & 0.160 \\
       \hline
RF    & $\tau_{tot}$    & 0.092 & 0.005 & $< 10^{-16}$ \\ 
      & $\tau_{\NDE}(1)$ & 0.153 & 0.003 & $< 10^{-16}$ \\ 
      & $\tau_{\NDE}(0)$ & 0.114 & 0.006 & $< 10^{-16}$ \\ 
      & $\tau_{\NIE}(1)$ & -0.022 & 0.003 & $5 \times 10^{-12}$ \\ 
      & $\tau_{\NIE}(0)$ & -0.060 & 0.003 & $< 10^{-16}$ \\ 
       \hline
GBM   & $\tau_{tot}$    & 0.157 & 0.004 & $< 10^{-16}$ \\ 
      & $\tau_{\NDE}(1)$ & 0.152 & 0.006 & $< 10^{-16}$ \\ 
      & $\tau_{\NDE}(0)$ & 0.146 & 0.004 & $< 10^{-16}$ \\ 
      & $\tau_{\NIE}(1)$ & 0.011 & 0.002 & $5 \times 10^{-6}$ \\ 
      & $\tau_{\NIE}(0)$ & 0.005 & 0.004 & 0.247 \\
\bottomrule[ 1pt]
\end{tabular}
\label{Table real3}
\end{table}

\begin{table*}[hbt!]
\centering
\caption{The validation losses of nuisance functions in real data application to income fairness.}

\begin{tabular}{lcccc}
\toprule[ 1pt]
  
                          & $\mathsf{DeepMed}$   & Lasso & RF    & GBM   \\
\midrule[ 1pt]
$a (1 | x, m)$         & 0.501 & 0.516 & 0.560 & 0.502 \\
$a (1 | x)$           & 0.600 & 0.612 & 0.631 & 0.600 \\
$\mu (x, 1, m)$         & 0.465 & 0.493 & 0.681 & 0.467 \\
$\E [\mu (X, D = 1, M) | X = x, D = 0]$ & 0.010 & 0.011 & 0.024 & 0.007 \\
$\mu (x, 1)$           & 0.479 & 0.510 & 1.040 & 0.480 \\
$\mu (x, 0, m)$         & 0.285 & 0.300 & 0.478 & 0.287 \\
$\E [\mu (X, D = 0, M) | X = x, D = 1]$ & 0.003 & 0.005 & 0.002 & 0.002 \\
$\mu (x, 0)$           & 0.288 & 0.306 & 0.711 & 0.291 \\

\bottomrule[ 1pt]
\end{tabular}
\label{Table real3:validation loss}
\end{table*}

\newpage
\bibliographystyleSM{plainnat}
\bibliographySM{Master.bib}

\end{document}